\documentclass[sigplan,screen,10pt]{acmart}
\usepackage{amsmath,amsfonts}
\usepackage[ruled,linesnumbered]{algorithm2e}
\usepackage{graphicx}
\usepackage{tabularx}
\usepackage{multirow}
\usepackage{multicol}
\usepackage{xcolor}
\usepackage{lipsum}
\usepackage{booktabs}
\usepackage{pifont}
\usepackage{xkeyval}
\usepackage{array}
\usepackage{subcaption}
\newcommand{\tabincell}[2]{\begin{tabular}{@{}#1@{}}#2\end{tabular}}
\usepackage{subcaption}
\usepackage{microtype}
\usepackage{mathtools}

\usepackage{booktabs}   
\usepackage{subcaption} 

\setcopyright{cc}
\setcctype{by}
\acmDOI{10.1145/3774934.3786434}
\acmYear{2026}
\copyrightyear{2026}
\acmISBN{979-8-4007-2310-0/2026/01}
\acmConference[PPoPP '26]{Proceedings of the 31st ACM SIGPLAN Annual Symposium on Principles and Practice of Parallel Programming}{January 31 -- February 4, 2026}{Sydney, NSW, Australia}
\acmBooktitle{Proceedings of the 31st ACM SIGPLAN Annual Symposium on Principles and Practice of Parallel Programming (PPoPP '26), January 31 -- February 4, 2026, Sydney, NSW, Australia}
\received{2025-09-01}
\received[accepted]{2025-11-10}

\begin{document}

\title{Accelerating Sparse Transformer Inference on GPU}

\author{Wenhao Dai}
\affiliation{
  \department{SSSLab, Dept. of CST}
  \institution{China University of Petroleum-Beijing} 
  \city{Beijing}
  \country{China}                    
}
\email{wenhao.dai@student.cup.edu.cn}          

\author{Haodong Deng}
\affiliation{
  \department{SSSLab, Dept. of CST}
  \institution{China University of Petroleum-Beijing} 
  \city{Beijing}
  \country{China}                    
}
\email{haodong.deng@student.cup.edu.cn}          

\author{Mengfei Rong}
\affiliation{
  \department{SSSLab, Dept. of CST}
  \institution{China University of Petroleum-Beijing} 
  \city{Beijing}
  \country{China}                    
}
\email{mengfei.rong@student.cup.edu.cn}          

\author{Xinyu Yang}
\affiliation{
  \department{School of Computer Science and Engineering}
  \institution{Beihang University} 
  \city{Beijing}
  \country{China}                    
}
\email{ltyxy@buaa.edu.cn}          

\author{Hongyu Liu}
\affiliation{
  \department{} 
  \institution{Baidu Inc.} 
  \city{Beijing}
  \country{China}                    
}
\email{liuhongyu02@baidu.com}          

\author{Fangxin Liu}
\affiliation{
  \department{School of Computer Science}
  \institution{Shanghai Jiao Tong University} 
  \city{Shanghai}
  \country{China}
}

\email{liufangxin@sjtu.edu.cn}          

\author{Hailong Yang}
\affiliation{
  \department{School of Computer Science and Engineering}   
  \institution{Beihang University} 
  \city{Beijing}
  \country{China}
}
\email{hailong.yang@buaa.edu.cn}          

\author{Qianwen Cao}
\affiliation{
  \department{College of Safety and Ocean Engineering}   
  \institution{China University of Petroleum-Beijing} 
  \city{Beijing}
  \country{China}                    
}
\email{qwcao@cup.edu.cn}          

\author{Qingxiao Sun}
\authornote{Corresponding author} 
\affiliation{%
 \institution{China University of Petroleum-Beijing}
 \institution{Beihang University}
 \city{Beijing}
 \country{China}}
\email{qingxiaosun@buaa.edu.cn}

\renewcommand{\shortauthors}{Dai et al.}

\begin{abstract}
Large language models (LLMs) are popular around the world due to their powerful understanding capabilities. As the core component of LLMs, accelerating Transformer through parallelization has gradually become a hot research topic. Mask layers introduce sparsity into Transformer to reduce calculations. However, previous works rarely focus on the performance optimization of sparse Transformer. In addition, current static operator fusion schemes fail to adapt to diverse application scenarios.
To address the above problems, we propose STOF, a framework that incorporates optimizations for Sparse Transformer that enables flexible masking and Operator Fusion on GPU. 
For multi-head attention (MHA) structure, STOF maps the computation to row-wise or block-wise kernels with unique storage formats according to analytical modeling.
For downstream operators, STOF maps the fusion scheme to compilation templates and determines the optimal running configuration through two-stage searching. 
The experimental results show that compared to the state-of-the-art work, STOF achieves maximum speedups of 1.6$\times$ in MHA computation and 1.4$\times$ in end-to-end inference.

\end{abstract}



\maketitle

\section{Introduction}
\label{sec:introduction}



Large language models (LLMs) have attracted widespread attention from industry and academia around the world~\cite{chang2024survey,achiam2023gpt,guo2025deepseek}. The massive parameters enable LLMs to capture the subtleties of human language~\cite{ouyang2022training}. In addition to general understanding, 
Transformer is the foundation of LLMs and the core of its powerful capabilities~\cite{zhao2023survey}. A variety of neural networks~\cite{devlin2019bert,radford2019language,Raffel2020T5} have evolved based on Transformer, while still retaining its encoding or decoding structure. The tensor operations involved in Transformer have rich parallelism, making it suitable for execution on many-core processors such as GPUs~\cite{fang2021turbotransformers}. This forces the performance optimization of Transformer for GPU architectures to become an important issue, which can bring huge economic benefits~\cite{aminabadi2022deepspeed}. 


Multi-head attention (MHA) is the essential building block in the Transformer model, where the attention module calculates the correlation among tokens in the input sequence~\cite{vaswani2017A}. The high-performance implementation of MHA fuses all tensor operations into one kernel, efficiently utilizing the memory hierarchy and function units~\cite{dao2022flashattention,zhai2023bytetransformer}. The novel MHA variants introduce mask layers to reduce the computational volume while maintaining accuracy~\cite{child2019generating}. The mask layer introduces sparsity to Transformer, and fragmented computation exacerbates the memory bandwidth bottleneck~\cite{wang2024raptor}. Furthermore, the explosive growth of masking patterns~\cite{zaheer2020big,beltagy2020longformer} makes it impractical to manually optimize each MHA variant separately. Although recent approaches~\cite{wang2024flashmask,dong2024flex} have supported a broader range of masking patterns with sparse representation or score modification, they are limited to continuous element distribution or suboptimal performance.


There are still potential optimization opportunities for downstream operators of MHA. Compilation-based operator fusion is adopted to reduce kernel launches and frequent I/O operations~\cite{zheng2020ansor,ma2020rammer}. DL frameworks~\cite{zheng2022astitch,ansel2024pytorch} generally only fuse memory-intensive (MI) operators, while compute-intensive (CI) operators are handled separately using vendor libraries. Other studies~\cite{niu2021dnnfusion,zheng2021neoflow,shi2023welder} have further explored the fusion of CI operator and MI operator to complement resource utilization such as memory bandwidth and streaming processors. The latest works~\cite{zheng2023chimera,zhang2024mcfuser} focus on the fusion of CI operators and improve performance in small-scale tensor computation with short sequences. However, the above rule-driven operator fusion schemes cannot adapt to diverse model hyperparameters and sequence lengths.




From the above analysis, sparse Transformer optimization faces the following challenges: 1) efficient kernel implementations with flexible representation of masking patterns; 2) adaptive operator fusion with sustained high performance for various computation scales; 3) fast exploration of hierarchical search space with fusion schemes and kernel parameters. 
We propose the STOF framework, which optimizes sparse Transformer inference through customized MHA kernels and adaptive operator fusion. STOF first determines the kernel implementation for MHA computation according to mask sparsity and sequence length. Then, STOF uses the encoding representation to specify the fusion scheme and maps it to compilation templates. Finally, STOF gradually expands the fusion range and determines the optimal scheme and its parameter setting via two-stage searching.

To the best of our knowledge, STOF is the first system to enable both flexible masking patterns and diverse operator fusion schemes for sparse Transformer scenarios. Specifically, STOF integrates  hand-tuned MHA kernels with generative compilation templates, providing a complete stack that establishes broader optimization opportunities.
We have selected typical networks with encoding or decoding structures including BERT~\cite{devlin2019bert}, GPT~\cite{radford2019language}, LLaMA~\cite{touvron2023llama}, ViT~\cite{dosovitskiy2020image}, and T5~\cite{Raffel2020T5} to verify the effectiveness of STOF. This paper makes the following contributions\footnote{The artifact for this paper is publicly available on Zenodo under DOI~\cite{dai2025STOF-AE}.}:





\begin{itemize}
\item We comprehensively analyze the impact of different masking patterns and inference configurations to expose potential optimization opportunities.
\item We propose a unified MHA module that implements row-wise and block-wise kernels with unique storage formats and optimizations. Besides, an analytical model is designed to determine kernel selection.
\item We propose an operator fusion module that converts the fusion scheme into compilation templates via numerical decoding. The search engine processes the encoded numerical representation and expands the fusion range based on performance feedback.


\item We develop an inference framework STOF that enables flexible masking patterns and determines the optimal operator fusion setting on GPU. The experimental results show that STOF achieves maximum speedups of 1.6$\times$ in MHA computation and 1.4$\times$ in end-to-end inference compared to the state-of-the-art work.

\end{itemize}

\section{Background}
\label{sec:background}

\subsection{Sparsity in Transformer Models} 
\label{sec:transformer_sparsity}

\subsubsection{Transformer Structure} Transformer~\cite{vaswani2017A} is widely recognized, where each encoder or decoder contains multiple multi-head attention (MHA) layers. The key operation of the MHA layer is scaled dot product attention (SDPA), which calculates the dot product of $Q$ and $K$, scales the result, optionally applies a mask at this stage, then applies the \texttt{Softmax} function to obtain the probabilities ($P$) and finally calculates the dot product of $P$ and $V$. Beyond MHA, Transformer includes downstream components: \texttt{Add} retains non-linear transformation information, \texttt{Norm} mitigates internal covariate shift via mean/variance normalization, and the \texttt{Feed Forward} layer comprises chained general matrix multiply (GEMM) operations with activations like \texttt{GELU} or \texttt{ReLU}. These components enable Transformer to handle complex cross-domain tasks while introducing operator characteristics that facilitate fusion-based optimizations.

\begin{figure}[htpb]
  \centering
  \includegraphics[width=\linewidth]{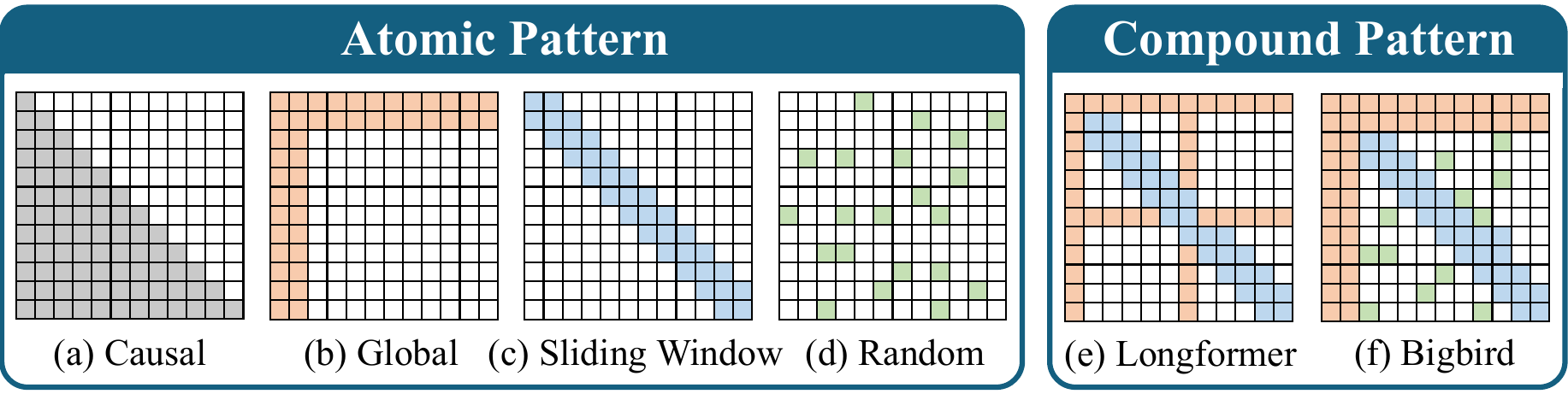}
  \caption{Atomic and compound sparse attention patterns.}
  \label{fig:bg-mask}
\end{figure}

\subsubsection{Sparse Attention Patterns}

Atomic sparse attention patterns are the building blocks of current popular sparse attention modules~\cite{adnan2024keyformer, liu2023deja, lin2022survey, roy2021efficient, beltagy2020longformer, zaheer2020big, child2019generating}. Figure~\ref{fig:bg-mask} (a)-(d) depict four most common atomic sparse attention patterns. The details are as follows.

\begin{itemize}
  \item \textit{Causal Attention.} To maintain temporal order, the query can access only preceding information, restricting connections to earlier nodes (the colored triangular).
    
  \item \textit{Global Attention.} Certain ``global'' nodes serve as central hubs, which receive information from others (the colored rows) and send it back (the colored columns).

  
  \item \textit{Sliding Window Attention.} Considering the concept of locality, the query only focuses on the neighboring nodes within a defined window size, with its mask matrix presenting a banded pattern (the colored bands).
  
  \item \textit{Random Attention.} The query block is randomly associated with the preceding and following information. By adjusting the filling rate, it has the possibility to discover accidental correlations (the colored blocks).
\end{itemize}

\subsection{Fused Kernel for MHA Structure}
\label{sec:kfam}

\begin{figure}[htpb]
  \centering
  \includegraphics[width=\linewidth]{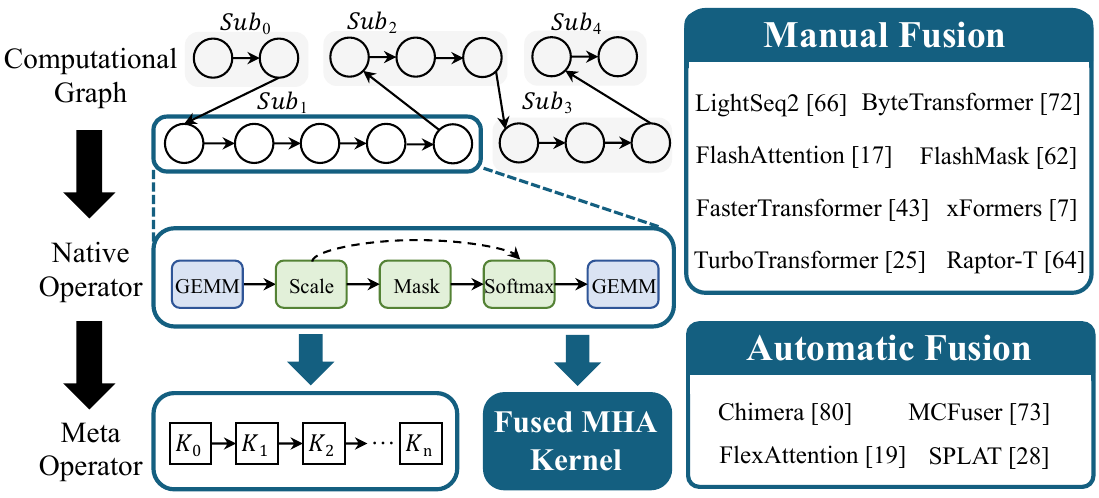}
  \caption{Kernel fusion for MHA computation.}
  \label{fig:bg-fusion}
\end{figure}

Numerous works~\cite{dong2024flex, wang2024flashmask,wang2024raptor,zhai2023bytetransformer,dao2022flashattention,wang2022lightseq2, fang2021turbotransformers,zheng2023chimera,zhang2024mcfuser,xFormers2022,nvidia22faster} have explored fusing MHA on GPU. Figure~\ref{fig:bg-fusion} shows a typical workflow of MHA fusion. The DL framework firstly parses the computational graph and captures the MHA sub-graph composed of coarse-grained native operators. Then, MHA fusion can be achieved manually or automatically. However, if the fusion of MHA with a certain mask layer is not supported, the sub-graph will be split into fine-grained meta operators to discover small-scale fusion opportunities.

Early works focus on the manual fusion of dense attention without the mask layer. 
ByteTransformer~\cite{zhai2023bytetransformer} adopts hand-written kernels: short sequences store the intermediate matrix entirely in shared memory (SMEM) and registers; longer sequences employ grouped GEMM to ease resource constraint. The customized kernels limit ByteTransformer to a maximum sequence length of 1,024. FlashAttention (FA) series\footnote{FA3~\cite{shah2024flashattention3} is only for GPUs with Hopper architecture and later.} becomes the most typical open source implementation.
FA~\cite{dao2022flashattention} partitions the input into blocks and passes the blocks to SMEM multiple times, gradually performing \texttt{Softmax} reduction. FA2~\cite{dao2023flashattention2} further partitions the work between warps within one block of attention computation to reduce the read and write of SMEM. However, FA only supports common masking patterns such as causal and sliding window. FlashMask~\cite{wang2024flashmask} extends FA with column-wise representation to exploit attention sparsity for skipped computations, integrated into PaddlePaddle~\cite{ma2019paddlepaddle} but unable to represent discrete distributions such as random attention.

For automatic fusion, the captured MHA sub-graph undergoes multi-level intermediate representation (IR) with hardware-independent (e.g., constant folding) and hardware-dependent (e.g., instruction scheduling) optimizations. MCFuser~\cite{zhang2024mcfuser} and Chimera~\cite{zheng2023chimera} accelerate MHA via GEMM chain loop scheduling but ignore hardware details like bank conflicts, performing poorly for long sequences. 
FlexAttention~\cite{dong2024flex} supports arbitrary masks by combining block masks with expression-based descriptions, but it is still constrained to fixed optimizations and achieves sub-optimal performance. 
SPLAT~\cite{gupta2025splat} focuses on bridging the performance gap of regular sparse kernels (R-SDDMM and R-SpMM) under structured sparsity (10\%–50\% non-zeros), yet this approach forgoes the opportunity to optimize MHA as a whole kernel.

\subsection{Hierarchical Space Exploration}
\label{sec:hse}

The hierarchical framework introduces a huge optimization space, making manual optimization on a case-by-case basis unrealistic. DL compilers~\cite{chen2018tvm,zhao2021akg,tillet2019triton} automatically explore opportunities across operator and kernel levels, deploying tensor programs on target hardware via IR conversion.

\subsubsection{Operator Fusion Opportunities} DL compilers predefine fusion rules that apply only to specific combinations, severely limiting the optimization space.
Researchers further classify tensor operators into MI and CI categories for selective fusion. Early works~\cite{zheng2022astitch,ansel2024pytorch} treat CI operators as non-fusion boundaries, fusing only MI operators to reduce off-chip accesses. Others~\cite{niu2021dnnfusion,shi2023welder} merge the CI operator with adjacent MI operators to balance hardware resource usage. Recent works~\cite{zheng2023chimera,zhang2024mcfuser} explore fusing CI chains by decomposing operators into blocks to break dependencies. However, due to GPU resource constraints, we notice that CI chain fusion only benefits on small scales.
Moreover, operator categories may shift with tensor dimensions, making category-based fusion schemes potentially suboptimal.

\subsubsection{Search Space Construction}
When fine-tuning the performance of DL models, the search space can be constructed by loop-based or template-based methods. The loop-based methods~\cite{zheng2020ansor,li2023exploiting} represent operators as deeply nested loops and optimize the statement execution via loop scheduling. Although hardware-universal, they lag vendor libraries due to ignoring hardware-specific instructions. The template-based methods~\cite{zheng2020flextensor,xing2022bolt,xu2023alt,chen2024evt} evolve as a new trend, which uses template primitives as building blocks to assemble complete DL models. The template primitives can map tensor programs to special function units like tensor cores. With hardware knowledge-driven tuning, they match vendor library performance. Bolt~\cite{xing2022bolt} derives primitives from CUTLASS~\cite{nvidia22cutlass} to support common fused operators. Due to the complex kernel structure of CUTLASS, further expanding the fusion range is too demanding for programmers.

\subsubsection{Auto-tuning Techniques} 



For loop-based construction, rule-based pruning first suppresses search space explosion, yet still amounts of configurations persist. Machine learning-driven cost models are trained online~\cite{zheng2020ansor} or offline~\cite{zheng2021tenset} to predict performance, integrated into heuristic searches (e.g., genetic algorithms) to speed up convergence. However, they all require sufficient runtime statistics. Aggressive techniques~\cite{ragan2013halide,ansel2024pytorch} unfold the computation graph sequentially, reducing search ranges from product to sum of operator spaces. But individual tuning without graph context leads to global suboptimal decisions.
In contrast, template-based construction maintains a constrained space aided by analytical models~\cite{kao2023flat,li2024accelerated} considering hardware and program details. Nevertheless, changes in the search space caused by operator fusion expansion remains unsolved.


We summarize comparisons of representative works and STOF in Table~\ref{tab:fused}. We implement compilation templates via the hardware abstraction of Triton~\cite{tillet2019triton} and TileLang~\cite{cheng2025pipethreader}. Both of them offer high-level programming interfaces that facilitate the template derivation for a wider fusion range. Then, the two-stage procedure encapsulating the AutoTune module quickly determines high-performance configurations.

\begin{table}[htbp]
\renewcommand{\arraystretch}{1.2}
\centering
\tiny
\caption{Comparison of representative works with STOF.}
\begin{tabular}{|l||ll|lll|}
\hline
\multirow{2}{*}{Name} & \multicolumn{2}{c|}{Operator Fusion}            & \multicolumn{3}{c|}{Hierarchical Search Space}                                                      \\ \cline{2-6} 
                        & \multicolumn{1}{l|}{Category}       & Expansion & \multicolumn{1}{l|}{Construction}   & \multicolumn{1}{l|}{Pruning}          & Searching             \\ \hline \hline
AStitch~\cite{zheng2022astitch}                 & \multicolumn{1}{l|}{MI-MI}          & Yes       & \multicolumn{1}{l|}{Rule}     & \multicolumn{1}{l|}{No}               & Breadth-First         \\ \hline 
Welder~\cite{shi2023welder}                  & \multicolumn{1}{l|}{CI-MI} & Yes       & \multicolumn{1}{l|}{Loop}     & \multicolumn{1}{l|}{No}               & Cost Model            \\ \hline
Chimera~\cite{zheng2023chimera}                 & \multicolumn{1}{l|}{CI-CI}          & No        & \multicolumn{1}{l|}{Loop}     & \multicolumn{1}{l|}{No}               & Analytical      \\ \hline
MCFuser~\cite{zhang2024mcfuser}                 & \multicolumn{1}{l|}{CI-CI}          & No        & \multicolumn{1}{l|}{Loop}     & \multicolumn{1}{l|}{Rule}       & Analytical      \\ \hline
Bolt~\cite{xing2022bolt}                    & \multicolumn{1}{l|}{General}      & No        & \multicolumn{1}{l|}{Template} & \multicolumn{1}{l|}{No}               & Analytical      \\ \hline
STOF (ours)             & \multicolumn{1}{l|}{General}      & Yes       & \multicolumn{1}{l|}{Template} & \multicolumn{1}{l|}{Analytical} & Reward-based \\ \hline
\end{tabular}
\label{tab:fused}
\end{table}

\section{Motivation}
\label{sec:motivation}

\begin{figure*}[htbp]
  \centering
  \begin{subfigure}[b]{0.49\linewidth}
    \centering
    \includegraphics[width=\linewidth]{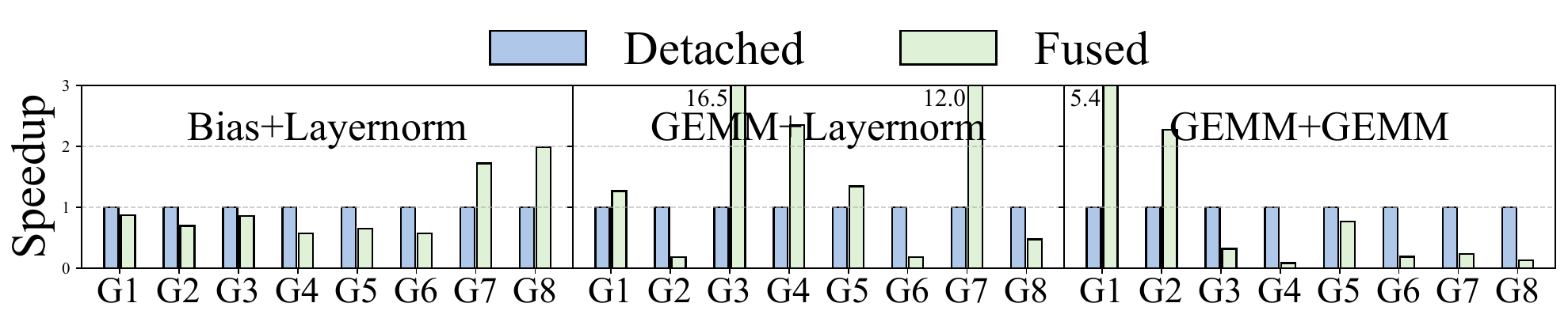}
    \caption{NVIDIA RTX 4090 GPU}
    \label{fig:detached}
  \end{subfigure}
  \begin{subfigure}[b]{0.49\linewidth}
    \centering
    \includegraphics[width=\linewidth]{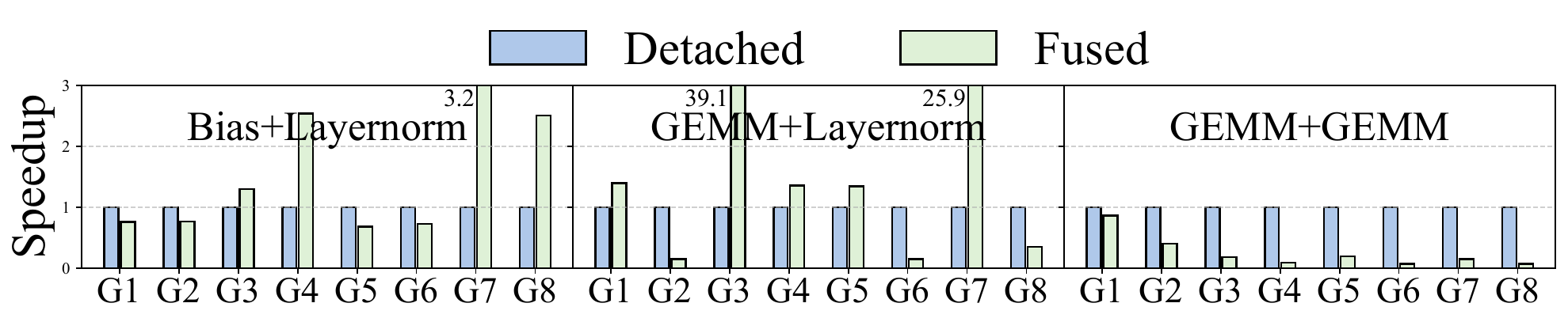}
    \caption{NVIDIA A100 GPU}
    \label{fig:fused}
  \end{subfigure}
  
  \caption{Performance comparison of detached operators and fused operator under different configurations.}
  \label{fig:triton_operators}
\end{figure*}

\begin{figure*}[htbp]
  \centering
  \begin{subfigure}[b]{0.49\linewidth}
    \centering
    \includegraphics[width=\linewidth]{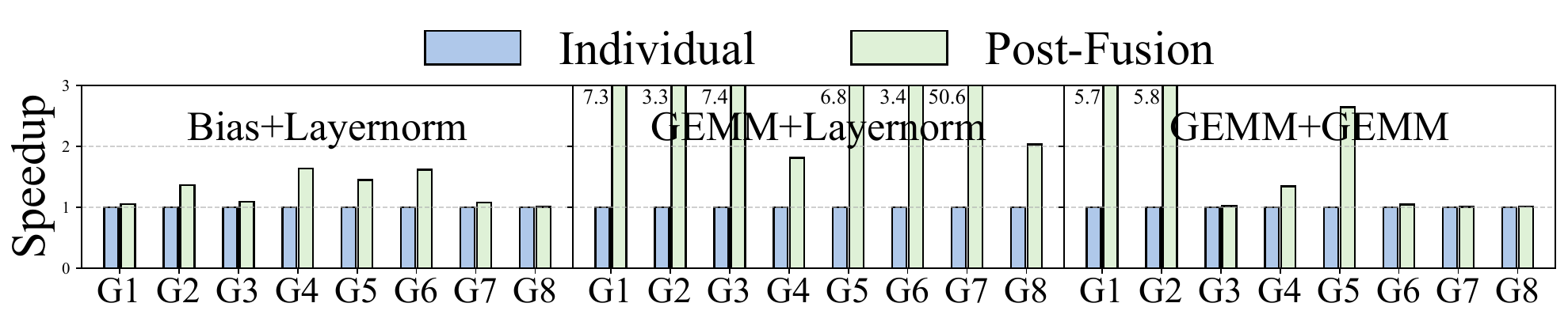}
    \caption{NVIDIA RTX 4090 GPU}
    \label{fig:single-config}
  \end{subfigure}
  \begin{subfigure}[b]{0.49\linewidth}
    \centering
    \includegraphics[width=\linewidth]{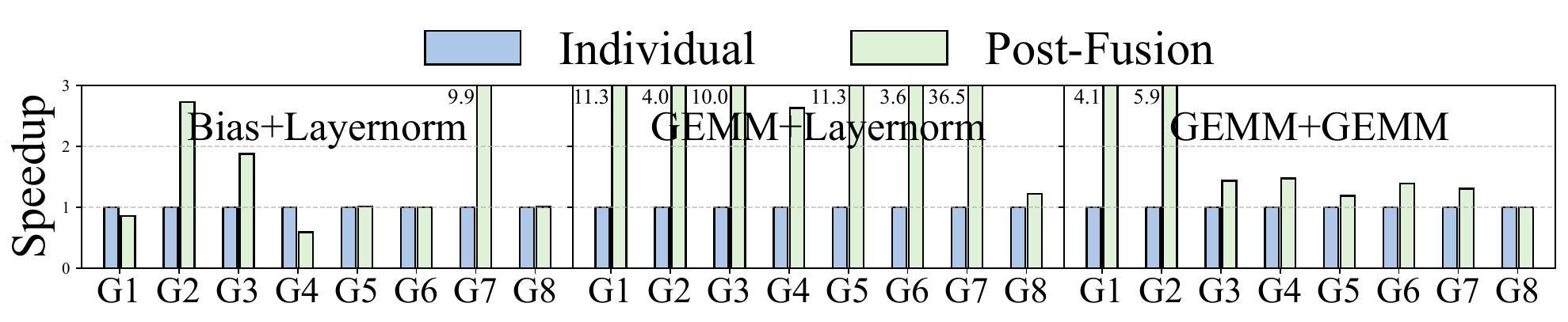}
    \caption{NVIDIA A100 GPU}
    \label{fig:fused-config}
  \end{subfigure}
  
  \caption{Performance comparison of fused operators using parameter settings from individual tuning and post-fusion tuning.}
  \label{fig:triton_configs}
\end{figure*}

\subsection{Diverse Features of Masking Patterns}
\label{sec:sparsity}

Within the MHA structure, sparse mask blocks part of the data elements, making it easier for the model to ``focus'' on the critical information. The mask layer is inserted between \texttt{GEMM} and \texttt{Softmax} operations, and the weights of the score matrix corresponding to the mask part are close to 0. Table~\ref{tab:moti_A_sparsity} lists the features of typical masking patterns with the sequence length ($seq\_len$) of 1,024. Consistent with previous works~\cite{child2019generating}, the band width and global width are set to $\sqrt{seq\_len}$ (i.e., 32). 
As seen, all masking patterns except the causal achieve a sparsity of over 80\%, while the sliding window even reaches 93.8\%. The above results provide optimization opportunities to skip useless computations.

\begin{table}[htbp]
\renewcommand{\arraystretch}{1.2}
\centering
\caption{Features of typical masking patterns.}
\tiny
\begin{tabular}{|l||ll|ll|ll|}
\hline
\multirow{2}{*}{\begin{tabular}[c]{@{}l@{}}Masking\\ Pattern\end{tabular}} & \multicolumn{2}{l|}{\multirow{2}{*}{\begin{tabular}[c]{@{}l@{}}Masking \\ Parameters\end{tabular}}}                    & \multicolumn{2}{c|}{Element Distribution}  & \multicolumn{2}{c|}{Sparsity}    \\ \cline{4-7} 
     & \multicolumn{2}{l|}{}    & \multicolumn{1}{l|}{Row}        & Column     & \multicolumn{1}{l|}{Type}    & Ratio  \\ \hline \hline
Causal                                                                    &  \multicolumn{2}{l|}{--}                & \multicolumn{1}{l|}{Continuous}   & Continuous   & \multicolumn{1}{l|}{Structured} & 50.0\% \\ \hline
     
\tabincell{l}{Sliding\\Window}                                                             & \multicolumn{2}{l|}{band width = 32}                                                                                   & \multicolumn{1}{l|}{Continuous} & Continuous & \multicolumn{1}{l|}{Structured}   & 93.8\% \\ \hline
Longformer                                                                 & \multicolumn{2}{l|}{\begin{tabular}[c]{@{}l@{}}global width = 32\\ band width = 32\end{tabular}}                       & \multicolumn{1}{l|}{Discrete}   & Discrete   & \multicolumn{1}{l|}{Structured}   & 88.8\% \\ \hline
Bigbird                                                                    & \multicolumn{2}{l|}{\begin{tabular}[c]{@{}l@{}}global width = 32\\ band width = 32\\ filling rate = 10\%\end{tabular}} & \multicolumn{1}{l|}{Discrete}   & Discrete   & \multicolumn{1}{l|}{Unstructured} & 80.8\% \\ \hline
\end{tabular}
\label{tab:moti_A_sparsity}
\end{table}

It is difficult for a data structure to represent sparsity features of various masking patterns. To achieve high kernel efficiency, FlashMask~\cite{wang2024flashmask} only supports the cases where the valid elements on the columns are continuous. This is because its data structure consists of four arrays that represent the start and end of two skipped regions. However, the discrete distribution of valid elements involves more skipped regions that cannot be represented. Bigbird integrates random patterns with unstructured sparsity, further complicating the mask representation. For unsupported masking patterns, previous works~\cite{zhai2023bytetransformer,dong2024flex} fall back to resetting the score matrix by subtraction after \texttt{GEMM}. This approach fails to jointly optimize \texttt{GEMM} and \texttt{Softmax} operations in the fused kernel.

\subsection{Potential Fusion Opportunities}
\label{sec:atuo-fusion}

Transformer structure still remains opportunities for operator fusion unexplored. If we roughly identify the operator types as MI or CI, the operator mixes can be enumerated into three categories. We fuse the operators of Transformer to evaluate the performance, where \texttt{Bias}+\texttt{Layernorm}, \texttt{GEMM+Layernorm}, and \texttt{GEMM+GEMM} represent MI+MI, CI+MI, and CI+CI mixes, respectively. Figure~\ref{fig:triton_operators} shows the speedup of the fused operator over the detached operators on NVIDIA RTX 4090 and A100 GPUs, where the x-axis represents the running configurations (detailed in Table~\ref{tab:moti_B_Config}).


\begin{table}[htbp]
  \small
  \centering
  \caption{The running configurations of fused operators.}
  \begin{tabular}{c || c | c | c }
    \hline
    Name & Batch Size & Sequence Length & Hidden Dimension \\
    \hline   \hline
    G1/G2 & 1 & 128 & 512/1024 \\  \hline
    G3/G4 & 1 & 4096 & 512/1024 \\ \hline
    G5/G6 & 8 & 128 & 512/1024 \\  \hline
    G7/G8 & 8 & 4096 & 512/1024 \\ \hline
  \end{tabular}%
\label{tab:moti_B_Config}
\end{table}

It can be observed that the effect of operator fusion varies significantly under different cases. For example, the fused \texttt{GEMM}+\texttt{Layernorm} operator achieves a maximum speedup of 16.5$\times$ and 39.1$\times$ when the hidden dimension is 512. But when the hidden dimension is 1,024, it results in significant slowdowns in most cases. The fused \texttt{GEMM}+\texttt{GEMM} operator achieves more than 2$\times$ speedup on RTX 4090 GPU when batch size and sequence length are 1 and 128, whereas it is inferior to the detached operators under all cases on A100 GPU. The above results indicate that fixed operator fusion schemes cannot adapt to diverse inference scenarios.

\subsection{Challenges in Parameter Tuning}
\label{sec:cos}

The combination of fusion schemes and kernel parameters constructs a hierarchical optimization space, making parameter tuning challenging. This stems from two key insights: 1) the search space of individual operators differs fundamentally from that of the fused operator; 2) the optimal parameter settings for individual and fused operators are inherently distinct. Figure~\ref{fig:triton_configs} shows the speedup of fused operators using parameter settings from post-fusion tuning over those from individual tuning on NVIDIA RTX 4090 and A100 GPUs. The x-axis represents the experimental configuration consisting of batch size, sequence length and hidden dimension. As seen, directly applying the optimal setting of individual operators to their fused implementation often leads to suboptimal performance. For example, \texttt{Bias}+\texttt{Layernorm}, \texttt{GEMM}+\texttt{Layernorm}, and \texttt{GEMM}+\texttt{GEMM} mixes achieve an average speedup of 2.4$\times$, 10.1$\times$, and 2.2$\times$ on A100 GPU, respectively. The results indicate that operator-by-operator sequential tuning is not a viable solution. On the other hand, naive global tuning can be inefficient due to the inconsistent search space.

\section{Methodology} 
\label{sec:method}

\subsection{Design Overview}
\label{sec:overview}


We propose STOF, accelerating Sparse Transformer inference with flexible masking patterns and operator fusion schemes on GPU.
STOF consists of a \textit{unified MHA module} and an \textit{operator fusion module}. The unified MHA module integrates row-wise and block-wise kernels with different storage formats, each with unique optimizations. The operator fusion module is embodied as the interaction between the fusion scheme converter and the hierarchical search engine. 

\begin{figure}[htbp]
  \centering
  \includegraphics[width=\linewidth]{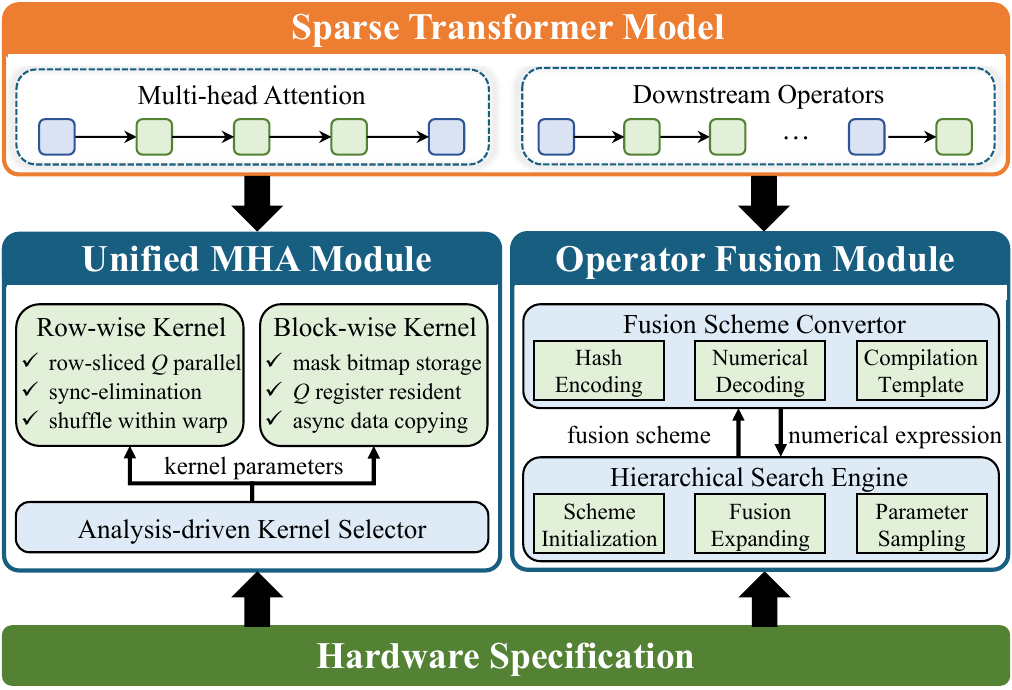}
  \caption{The design overview of STOF.}
  \label{fig:4-method-overview}
\end{figure}

Figure~\ref{fig:4-method-overview} illustrates the design overview of STOF. STOF divides the sparse Transformer model into MHA structure and downstream operators. This ensures both the customization of MHA and the flexibility of operator fusion. For MHA structure, STOF maps its calculations directly to GPU kernels with fine-grained optimization. The kernel selector determines the MHA kernel by applying an analytical model that takes hardware specifications into account. For downstream operators, the scheme converter expresses the fusion scheme as a binary array through hash coding upwards and maps it to compilation templates through numerical decoding downwards. The search engine initializes scheme, expands fusion, and samples parameters via analytical modeling, performance feedback, and reward algorithm, respectively.

We have implemented two sets of kernels depending on the data partitioning granularity. The row-wise kernel slices $Q$ into rows to achieve high locality. Moreover, the row-wise kernel applies shuffle within a warp and eliminates the synchronization among warps, improving performance at small input sizes. In contrast, the block-wise kernel is more general with fine-grained block partitioning, where $Q$, $K$, and $V$ are partitioned into sub-blocks and put into SMEM to utilize the GPU memory hierarchy. Since row partitioning can be regarded as an extreme case of block partitioning, we elaborate on the block-wise optimizations in Section~\ref{subsec:unified-mha}.

The main takeaway of STOF is a novel co-design that bridges manual kernel implementation for sparse MHA structure and automatic fusion for dense downstream operators. Specifically, the sparsity in STOF is exclusively handled within the MHA module, where mask-based computation is explicitly managed by customized kernels.
All subsequent operators after MHA are dense and executed via template-based fusion, ensuring both high performance and compositional flexibility. Beyond the specific optimizations for Transformer architectures, the core methodology of STOF is readily extensible to emerging LLM architectures. For instance, in Mixture-of-Experts (MoE) models~\cite{cai2025survey}, we can accelerate activated experts via specialized kernels while optimizing the routing logic through template-based fusion, potentially supporting dynamic computation paths at minimal cost.


\subsection{Unified MHA Kernels}
\label{subsec:unified-mha}
\subsubsection{Sparse Storage Format}

\begin{figure}[htbp]
  \centering
  \includegraphics[width=\linewidth]{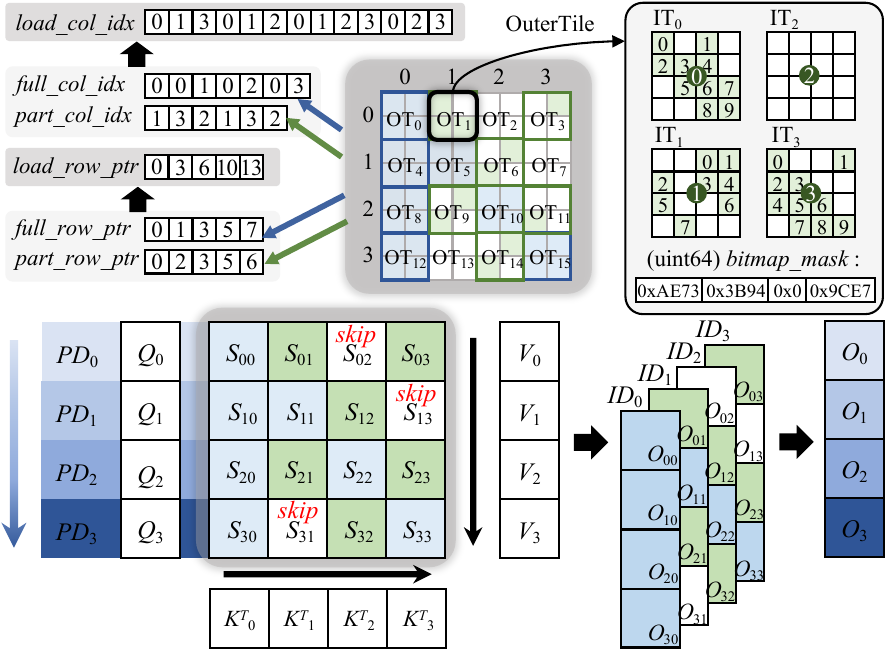}
  \caption{MHA computation with two-level storage format.}
  \label{fig:4-method-Attn-kernel}
\end{figure}

Figure~\ref{fig:4-method-Attn-kernel} shows the block-wise computation with a sparse storage format that can represent arbitrary mask. Inspired by literature~\cite{niu2022tilespgemm, fan2025spinfer}, we adopt a two-level storage format combining Block Compressed Sparse Row (BSR) and bitmap, preserving sparsity while enabling structured computation. As shown in Figure~\ref{fig:4-method-Attn-kernel}, we abstract two levels as OuterTile (OT) and InnerTile (IT) to reveal globally skipped blocks and intra-block element distribution, respectively. Each OT is composed of 64 8$\times$8 ITs (only 4 are shown in the figure for clarity). An OT is marked as ``full'' if all of its ITs are not empty, otherwise ``part''. For the ``full'' OTs, the difference between $full\_row\_ptr[i]$ and $full\_row\_ptr[i - 1]$ indicates the number of ``full'' OTs in the $i$-th row. The array $full\_col\_idx$ specifies the column indices of ``full'' OTs. For example, as can be inferred from $full\_row\_ptr$ and $full\_col\_idx$ arrays in the figure, the column indices of ``full'' OTs in the $2$-nd row are 0 and 2.




For the  ``part'' OTs, there are also two similar arrays including $part\_row\_ptr$ and $part\_col\_idx$. The $part\_col\_idx$ further points to the corresponding IT with sparse element distribution.
Since each IT contains exactly 64 elements, it can be efficiently represented by a single \texttt{uint64} value. 
Consequently, for each ``part'' OT, the internal mask information is stored as a \texttt{bitmap\_mask} array consisting of 64 \texttt{uint64} elements.
During the processing of the innermost loop, each \texttt{bitmap\_mask[i]} is retrieved to obtain the precise masking pattern. By combining the structures of ``full'' and ``part'' OTs, we obtain $load\_row\_ptr$ and $load\_col\_idx$ arrays that directly specify the location of non-empty OTs in the mask.


\label{sec:fmk}

\subsubsection{Kernel Implementation}
We cut the input tensor $Q$ into sub-blocks of size (\textit{OT\_Size\_M}, \textit{head\_size}) along the $seq\_len$ dimension, as illustrated in Algorithm~\ref{alg:AttnKernel}.
Each sub-block $Q_i$ (line 2) corresponds to a Row-\underline{P}arallel \underline{D}imension ($PD_i$), where $i \in [0,\lceil\frac{seq\_len}{OT\_Size\_M}\rceil)$. 
To enhance data locality, for each row processed by $Q_i$, $K$ and $V$ are divided into sub-blocks $K^{T}_{j}$ and $V_{j}$ of size $(OT\_Size\_N, head\_size)$ (lines 7-9), where $j \in [0, \lceil\frac{seq\_len}{OT\_Size\_N}\rceil)$. The workload of OTs per row is determined by the arrays $load\_row\_ptr$ and $load\_num$ (lines 4-6). Under the coarse-grained block of size $({OT\_Size\_M},{OT\_Size\_N})$, only valid OTs that require computation are loaded, while others are skipped. This alleviates bandwidth conflicts by greatly reducing global memory access.
The asynchronous copy of $V_{j}$ (line 9) allows the \texttt{GEMM} (line 10) to proceed without waiting for the completion of $V_{j}$'s transfer. Furthermore, it eliminates the need for data loading stalls in the subsequent \texttt{GEMM} (line 16). After obtaining $P_{ij}$, the presence of any ``part'' OTs in the current row is checked to determine whether ITs' storage information should be loaded from the \texttt{uint64} array bitmap\_mask and applied to mask $S_{ij}$ (lines 11-14). Due to the consistency of $K^{T}_{j}$ and $V_{j}$ blocks on the \underline{I}teration \underline{D}imension ($ID_j$), the skip operation on $K^{T}_{j}$ is also applied to $V_{j}$, thus reducing amounts of calculation and storage. After the \texttt{Softmax} operation, $S_{ij}$ and the scaling factor $\alpha$ within the OT are obtained to ensure the correctness of reduction operations (lines 15-16). Finally, the results are written back to HBM (line 18).


\begin{algorithm}
    \SetAlgoLined
    \scriptsize
    \caption{MHA Kernel with Unified Format}
    \label{alg:AttnKernel}
    \KwIn{flattened tensors on HBM $Q\_HBM$, $K\_HBM$, $V\_HBM$; unified mask storage structures $part\_row\_ptr$, $part\_col\_idx$, $load\_row\_ptr$, $load\_col\_idx$, $bitmap\_mask$}
    \KwOut{MHA result on HBM $result\_HBM$}
    \BlankLine

    \For{$i$ in $[0$, $ \lceil\frac{seq\_len}{OT\_Size\_M}\rceil)$}{
        $Q_i \leftarrow$Load\_from\_HBM$(Q\_HBM_i)$\;   
        $tmp\_part\_col\_idx, O_{i}\leftarrow 0$\;
        $load\_num \leftarrow load\_row\_ptr[i+1] - load\_row\_ptr[i]$\;
        $part\_num \leftarrow part\_row\_ptr[i+1] - part\_row\_ptr[i]$\;
        \For{$kv\_idx$ in $[0, load\_num)$}{
            $j \leftarrow load\_col\_idx[load\_row\_ptr[i] + kv\_id]$\;
            $K^T_{j} \leftarrow $Load\_from\_HBM$(K\_HBM_{j})$\;
            $V_{j} \leftarrow$ \texttt{\_\_async\_memcpy}$($Load\_from\_HBM$(V\_HBM_{j}))$\;
            $P_{ij} \leftarrow$ Compute\_GEMM $(Q_i, K^{T}_{j})$\;
            \If{$tmp\_part\_col\_idx < part\_num$ and $j == part\_col\_idx[part\_row\_ptr[i] + tmp\_part\_idx]$}{
                Apply\_Mask$(S_{ij}, bitmap\_mask[tmp\_part\_col\_idx])$\;
                $tmp\_part\_col\_idx \leftarrow tmp\_part\_col\_idx + 1$\;
            }
            $S_{ij}, \alpha \leftarrow $Softmax$(P_{ij})$\;
            $O_{i} \leftarrow O_{i} \times \alpha + $Compute\_GEMM$(S_{ij}, V_{j})$\;
        }
        $result\_HBM \leftarrow $Write\_back\_to\_HBM$(O_{i})$.
    }

\end{algorithm}


We further conduct advanced optimizations on the MHA kernel, primarily based on FA2~\cite{dao2023flashattention2}. For example, the 8$\times$8 size of ITs not only matches the \texttt{uint64} size but also aligns with the data granularity operable by Tensor Cores. Notably, OTs are stored in row-major order to accommodate the row-wise iterative computation of \texttt{Softmax}, whereas ITs are stored in col-major order to enable bank conflict-free accesses. The OT size is determined by considering cache capacity and the number of SMs. 
During each iteration, $Q_{i}$ is kept in registers, $K^{T}_{j}$ and $V_{j}$ share a single physical portion of shared memory.




\subsubsection{Kernel Selection}
By comprehensively considering the influence of masking patterns and sequence lengths, we decide whether to apply a row-wise or block-wise kernel for MHA computation. As formulated in Equation~\ref{equ:kernel-seclector-1}, we empirically set the coefficient $\tau$ to 1.2 and calculate the $threshold$. We select row-wise kernel if $threshold$ is less than 0, indicating that the ratio of valid OTs (i.e., ``full'' and ``part'') is sufficiently low. Note that we use $log$ operation to penalize the extremely sparse situation due to the increase of $sep\_len$ while the mask width remains unchanged. By doing so, we have limited row-wise kernel to cases where the number of valid OTs is small and the $sep\_len$ is short. In such cases, centralized row-wise computation of mask elements brings excellent data locality. For other general cases, we apply block-wise kernel to maximize performance.



\begin{equation}
 \footnotesize
threshold = \frac{{load\_row\_ptr }[\lceil\frac{seq\_len}{16}\rceil]}{(\lceil\frac{seq\_len}{16}\rceil)^{2}} - \frac{\tau}{(\log_2\lceil\frac{seq\_len}{16}\rceil)^{2}}
\label{equ:kernel-seclector-1}
\end{equation}



\subsection{Fusion Scheme Conversion}
\label{sec:fse}

It is essential to express the fusion scheme appropriately, quantifying the dependencies among vertical operators and identifying the fusion boundaries. Inspired by the high-low voltage levels of digital circuits, we use binary hash codes as the numerical expression of fusion schemes. STOF maps the fused operators to compilation templates so that the compiler can further add kernel-level optimizations. From the perspective of the computational graph, the captured adjacent nodes are replaced with fused nodes.

\begin{figure}[htbp]
  \centering
  \includegraphics[width=\linewidth]{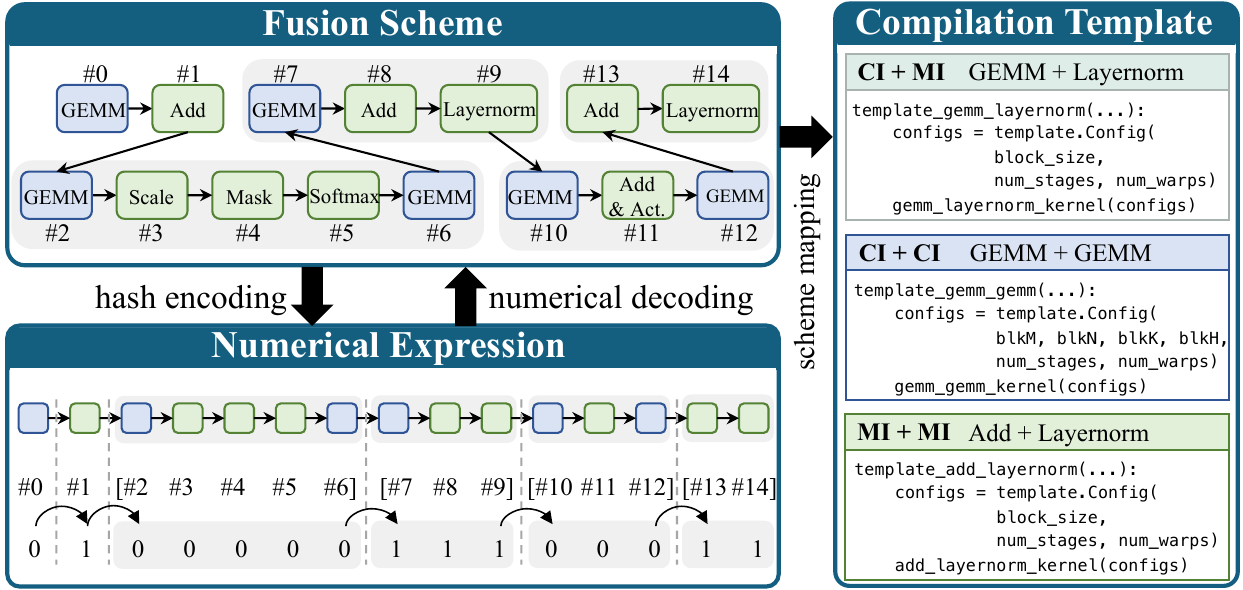}
  \caption{The workflow of fusion scheme converter.}
  \label{fig:4-method-encoding}
\end{figure}

Figure~\ref{fig:4-method-encoding} shows the workflow of the fusion scheme converter in STOF. Take the forward propagation of BERT as an example, STOF traverses the computational graph constructed by the DL framework and extracts subgraphs that conform to the patterns of fusion schemes. Each subgraph is mapped to the target compilation template, which is carefully implemented to achieve optimal performance. Specifically, the templates decompose tensor operations into tiles to maximize data reuse, leverage warp-level primitives for efficient reductions, and apply multi-stage pipelining to overlap memory accesses and computation.
Although we customize the compilation template according to the functionality of the fused operator, the graph mapping process is highly flexible. For instance, the template that computes a GEMM chain with CI+CI pattern can also incorporate simple MI operations, such as adding bias element by element (i.e., \texttt{Bias}). On the other hand, the compilation template hides the hardware execution details and only exposes key kernel parameters for performance tuning. For the GEMM chain, the sub-block sizes and the launch configuration (e.g., number of stages) constitute the search space, providing the possibility of further optimization targeting at a specified case.

The fusion scheme is quantized by hash encoding, and the native operators are represented as arrays with a length equal to the number of operators according to the vertical fusion situation. 
In this way, hash encoding translates abstract fusion patterns into a quantifiable space, a process that establishes a bidirectional mapping consistent with the definition of ``hash''.
We assume that in addition to mapping MHA ([\#2-\#6]) to the fused kernel, the fusion scheme also specifies three other downstream fused operators including [\#7-\#9], [\#10-\#12], and [\#13,\#14]. 
The numbers representing the operators in the subgraph are the same, which is similar to the high-low voltage levels of the circuit. 
For example, the numbers corresponding to the subgraph [\#7-\#9] are all 1. Besides, the different numbers of adjacent operators refer to the boundary of adjacent subgraphs. 
Note that the numbers are unrelated to the operator characteristics, they are introduced solely to facilitate the subsequent tuning process.
The numerical expression is usually in binary, but it can also be converted to hexadecimal format with a higher compression rate. Intuitively, this expression approach constructs a flexible search space that can represent any fusion scheme. On this basis, we propose a two-stage search mechanism to tune the running configuration during inference.

\subsection{Search Space Exploration}
\label{sec:psc}

\begin{figure}[htbp]
  \centering
  \includegraphics[width=\linewidth]{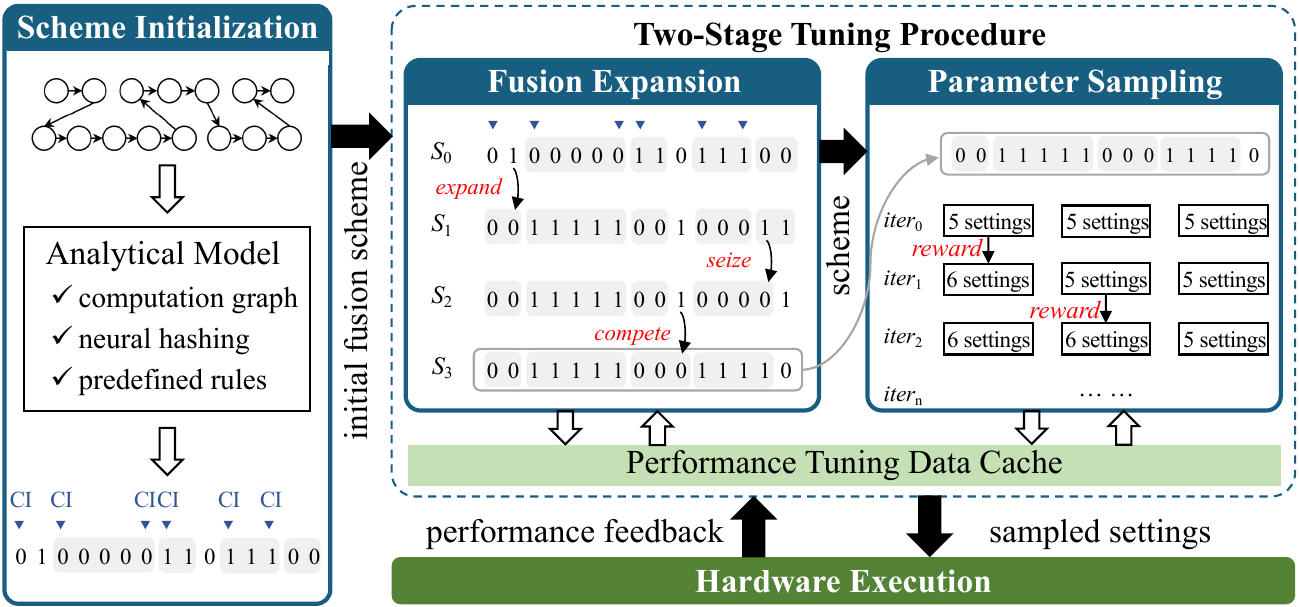}
  \caption{The workflow of hierarchical search engine.}
  \label{fig:4-method-search}
\end{figure}

STOF deploys a search engine featuring scalable fusion boundaries and parameter-tuning capabilities. As depicted in Figure~\ref{fig:4-method-search}, 
the search engine first uses neural hashing and predefined rules to derive an initial fusion scheme. Then, the two-stage procedure is conducted to determine the boundaries of the fused operators and their kernel parameters.


\subsubsection{Fusion Scheme Initialization}
STOF leverages both pattern discovery and expert knowledge to derive the initial fusion scheme.
First, STOF adopts a convolutional subgraph analysis method \textit{neural hashing} to discover representative subgraphs that frequently appear during the inference, formalized as: $H(G) = \mathcal{F}_{\text{hash}}(\mathcal{F}_{\text{conv}}(G))$. Here, $G$ is the input computational graph structure; $\mathcal{F}_{\text{conv}}$ is a convolutional feature extractor that extracts local structural features from the graph $G$. $\mathcal{F}_{\text{hash}}$ is a hash mapper, which compresses and discretizes the extracted features into a unique hash fingerprint $H(G)$. By analyzing the frequency distribution of these fingerprints, STOF can rapidly detect classical subgraph structures across Transformer-based models. Second, STOF uses \textit{predefined rules} to extract potentially high-performance subgraphs from the identified subgraph structures to form the initial scheme.
For example, according to the conclusion in Section~\ref{sec:motivation}, the GEMM chain is preferentially fused into one segment under smaller batch sizes and sequence lengths.




\subsubsection{Two-Stage Tuning Procedure}
In \textit{the first stage}, STOF tends to expand the boundaries of the segments until there is no additional benefit after fusion.
Since DL frameworks have implemented the fusion of common MI operators, we mark CI operators and adjust the fusion scheme around them for complementarity.
We have restricted that there are at most two CI operators in each segment, and classified the fusion rules into the following three categories.

\begin{itemize}
  \item \textit{expand}: merge existing individual or fused operators to form a new segment without disrupting the structure of other segments, such as the transition from $S_0$ to $S_1$.
  
  \item \textit{seize}: a segment with at least one CI operator preempts an operator from a segment consisting of only MI operators, such as the transition from $S_1$ to $S_2$.

  \item \textit{compete}: if two segments compete for an individual operator, the segment with only one CI operator will be extended first, such as the transition from $S_2$ to $S_3$.

\end{itemize}

Based on the above rules, we apply depth-first search (DFS) to gradually expand the fusion range. In this process, STOF randomly samples a fixed number of parameter settings of the pre-fusion and post-fusion operators, then takes the best setting to compare the performance. If there is a performance gain, STOF will keep the new fusion scheme, otherwise roll back. As long as the scheme has appeared and the performance under specific parameter settings is recorded in the cache, the same attempt will not be made later. 

In \textit{the second stage}, STOF conducts parameter sampling for the determined scheme. Specifically, we fix the total number of configurations during each iteration and retrieve performance data. In the first iteration, STOF ensures the number of sampled settings for each segment is the same. When the highest overall gain is achieved when tuning a segment, STOF rewards it by increasing the sampled settings in the next iteration. Similarly, STOF caches performance data to avoid repeated execution of the same parameter setting.

\begin{figure*}[htbp]
    \centering
    \includegraphics[width=\linewidth]{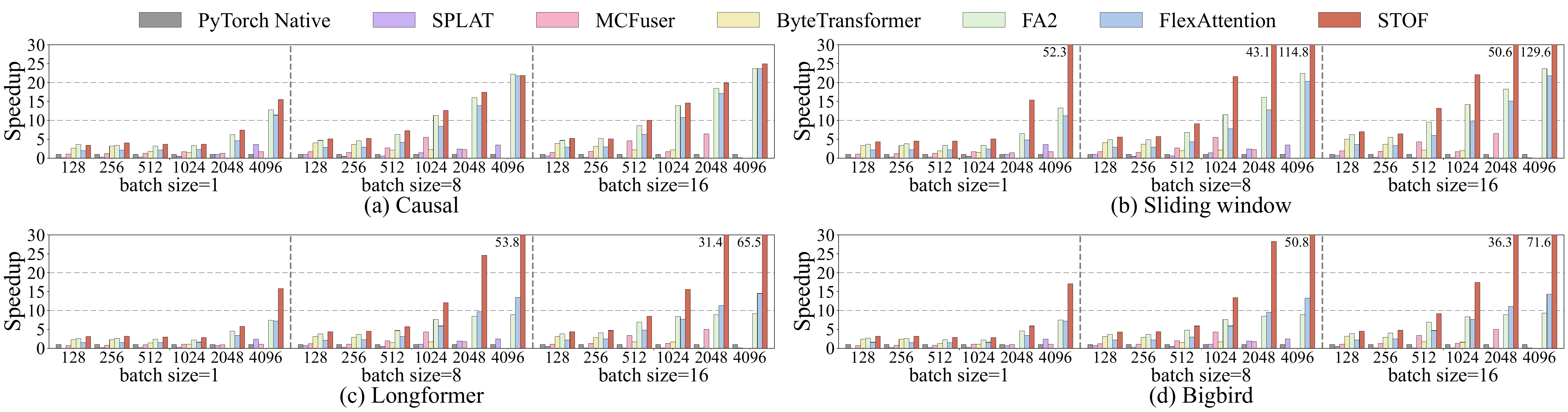}        
    \caption{The MHA performance of the methods normalized to that of PyTorch Native on NVIDIA RTX 4090 GPU.}
    \label{fig:MHA-evalaution-4090}
\end{figure*}

\begin{figure*}[htbp]
    \centering
    \includegraphics[width=\linewidth]{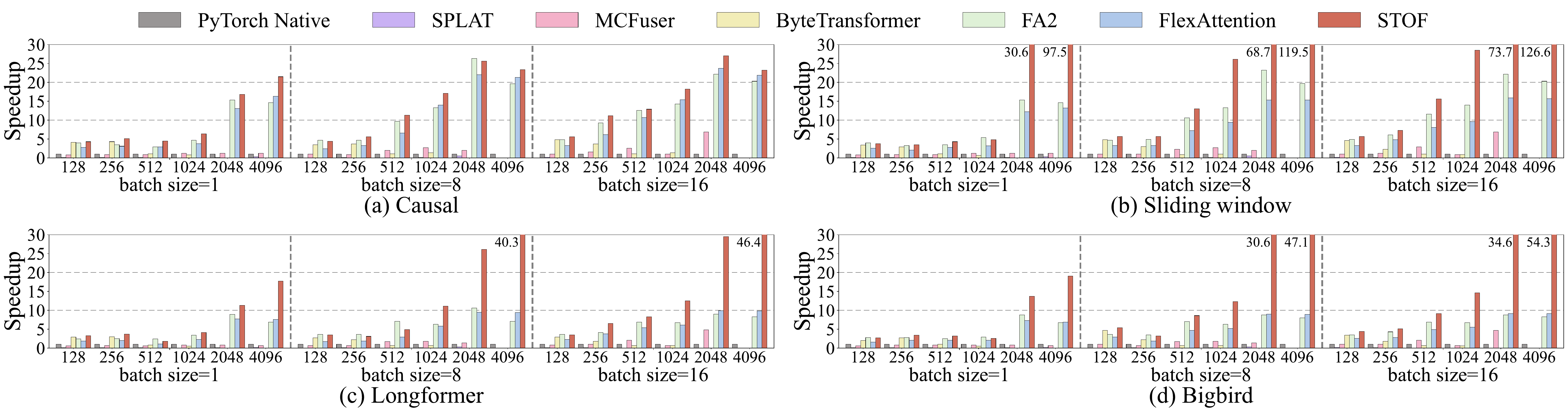}         
    \caption{The MHA performance of the methods normalized to that of PyTorch Native on NVIDIA A100 GPU.}
    \label{fig:MHA-evalaution-A100}
\end{figure*}

\subsection{Implementation Details}
\label{sec:impdetails}

We have implemented a system prototype of STOF based on PyTorch~\cite{ansel2024pytorch}, Triton~\cite{tillet2019triton} and TileLang~\cite{cheng2025pipethreader}, involving approximately 5,000 LOC of Python and 2,500 LOC of C/CUDA. 
The block-wise kernel is developed based on FA2~\cite{dao2023flashattention2} with the CuTe structure, but introduces an efficient two-level storage format and corresponding optimizations.
Subsequently, the customized MHA kernel is loaded into PyTorch through the \texttt{torch/cpp\_extension} interface, which encapsulates the kernel in the form of a native function.
When the MHA kernel is first called, it is just-in-time (JIT) compiled into a shared object file (\texttt{.so}) using the \texttt{ninja} tool, enabling dynamic linking at runtime without repeated compilation. 


Regarding the operator fusion module, we find that the Triton- and TileLang-based compilation templates demonstrate performance variance under different fused operators, so we select the implementation that achieves superior performance in each case.
We enable the graph capture and replacement by manipulating objects of type \texttt{fx.GraphModule}. 
Since the overall implementation of STOF is
compatible with the \texttt{torch.compile} function, its related compilation optimizations can be reused to maximize performance.

\section{Evaluation}  
\label{sec:evaluation}

\subsection{Experiment Setup} 
\label{sec:setup}

\subsubsection{Hardware and Software Platforms} 

We evaluate STOF on two generations of GPUs, including NVIDIA RTX 4090 of Ada model and NVIDIA A100 of Ampere model. The experiments are conducted in the software environment configured with Ubuntu 22.04, CUDA v12.6, and PyTorch 2.7.0. We package Docker containers to quickly migrate the software environment between hardware platforms.


\subsubsection{Comparison Configurations and Methods}

We conduct evaluation on both atomic and compound masking patterns including causal, sliding window, Longformer~\cite{beltagy2020longformer}, and Bigbird~\cite{zaheer2020big}. The sequence length ranges from 128 to 4,096 with a stride of 2$\times$, and the batch size ranges from 1 to 16. For MHA computation, we follow the configuration of BERT-Base. For end-to-end inference, the configuration is set to be consistent with the standard models of BERT~\cite{devlin2019bert}, GPT2~\cite{radford2019language}, LLaMA~\cite{touvron2023llama}, T5~\cite{Raffel2020T5} and ViT~\cite{dosovitskiy2021vit}. We compare STOF with PyTorch Native, PyTorch Compile~\cite{ansel2024pytorch}, FlashAttention2 (FA2)~\cite{dao2023flashattention2}, FlexAttention~\cite{dong2024flex}, ByteTransformer~\cite{zhai2023bytetransformer}, Bolt~\cite{xing2022bolt},  MCFuser~\cite{zhang2024mcfuser}, and SPLAT~\cite{gupta2025splat}. Note that FlexAttention, FA2, and SPLAT are optimized only for MHA, while PyTorch Compile integrates FA2 for MHA computation. In addition, Bolt has no MHA-specific optimizations and only appears in the end-to-end evaluation. Since SPLAT is not open source, we reproduce it based on the contents in the paper. We adopt the half precision (FP16) for evaluation, which is commonly used for model inference in industry~\cite{aminabadi2022deepspeed}, ensuring a unified comparison across all methods. To minimize machine errors, we perform warm-ups for all experiments and run 100 times to record the average performance.


\begin{figure*}[htbp]
    \centering
    \includegraphics[width=\linewidth]{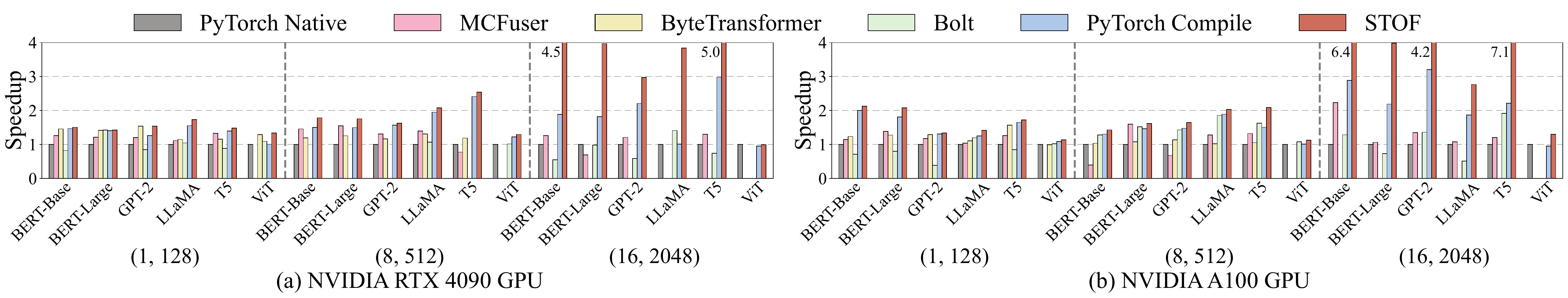}
    \caption{The end-to-end performance of the methods normalized to that of PyTorch Native on RTX 4090 and A100 GPUs.}
    \label{fig:eva-EE-4090-A100}
\end{figure*}

\begin{table*}[htbp]
\scriptsize
\caption{Tuning time of STOF, MCFuser, and Bolt for end-to-end inference on A100 GPU in seconds.}
\centering 
\renewcommand{\arraystretch}{1.2}
\setlength{\tabcolsep}{2.7pt} 
\begin{tabular}{|l||c|c|c|c|c|c||c|c|c|c|c|c||c|c|c|c|c|c|}
\hline
Input Size & \multicolumn{6}{c||}{(1, 128)}  & \multicolumn{6}{c||}{(8, 512)} & \multicolumn{6}{c|}{(16, 2048)} \\ \hline
Name &  BERT-B & BERT-L & GPT & LLaMA  & T5  & ViT  & BERT-B & BERT-L & GPT  & LLaMA  & T5  & ViT  & BERT-B & BERT-L & GPT   & LLaMA  & T5  & ViT \\ \hline \hline
MCFuser & 51.4   & 52.4   & 49.5 & 48.8 & 71.9 & 100.2  & 91.8   & 132.3   & 100.8 & 110.8 & 239.0 & 437.8  & 660.2   & 1049.7   & 664.4 & 820.6 & 1987.6 &  4264.3 \\ \hline
Bolt    & 53.3  & 57.3  & 48.8 &  52.1 & 70.7 &  120.7 & 90.8   & 126.1   & 99.8  & 124.6 & 244.7 &  468.8 & 652.2   & 1067.7   & 738.6 & 837.0 & 1860.8  & 3848.6 \\ \hline
STOF (ours)     & \textbf{23.3}   & \textbf{22.6}   & \textbf{23.8}  &  \textbf{29.5}  & \textbf{43.1}  &  \textbf{93.9}   & \textbf{40.9}   & \textbf{55.0}  & \textbf{40.9}  &  \textbf{43.6}  & \textbf{80.3} &  \textbf{99.3}  & \textbf{99.6}    & \textbf{225.3}     & \textbf{122.2} &  \textbf{264.6} & \textbf{388.3}   & \textbf{412.8} \\ \hline
\end{tabular}
\label{tab:tuning_time}
\end{table*}

\subsection{MHA Performance}
\label{sec:fusedMHA}

Figure~\ref{fig:MHA-evalaution-4090} and Figure~\ref{fig:MHA-evalaution-A100} present the MHA performance of the methods normalized to that of PyTorch Native on RTX 4090 and A100 GPUs. The missing bars are attributed to two reasons: 1) ByteTransformer lacks support for sequence lengths greater than 1,024; 2) MCFuser runs out of memory (OOM) when the input scale is large. As seen, STOF shows consistent superior performance on both GPU platforms.
Compared to the state-of-the-art FlexAttention implementation, STOF achieves the average speedups of 1.8$\times$ and 1.6$\times$ on RTX 4090 and A100 GPUs, respectively. 
STOF achieves superior performance on sliding window mask because its high sparsity and concentration of valid blocks facilitate computation skipping.
Even for causal masks, STOF still achieves a certain speedup over FA2 and FlexAttention under most cases. The reason is that the two-level storage format combining BSR and bitmap further improves on-chip memory locality. In contrast, due to the lack of tensor core support, SPLAT achieves decent performance on RTX 4090 GPU with higher CUDA core ratio, achieving a maximum speedup of 3.6$\times$ compared to PyTorch Native; but it lags behind on A100 GPU across all cases.

The above figures illustrate the MHA performance at different input scales in detail. At small scales, STOF achieves relatively better performance than FA2 and FlexAttention under most cases. STOF enables the row-wise kernel, where the use of shuffle operations within the warp incurs extremely low synchronization cost.
On the other hand, STOF achieves significant speedup compared to other methods at large input scales. For example, when the setting of (batch size, sequence length) is (16, 4,096), STOF achieves 4.8$\times$ and 4.9$\times$ speedups over FA2 and FlexAttention on A100 GPU, respectively. This is mainly because the block-wise kernel makes full use of the mask sparsity to skip unnecessary calculations. Besides, the optimizations such as asynchronous data copying and \textit{Q} register resident serve as the foundation for performance improvement.
Note that PyTorch Native, MCFuser, and ByteTransformer do not natively support sparse masks. The basic approach is to subtract the mask matrix, thus missing the opportunity to reduce the amount of calculation.

\subsection{End-to-end Performance}
\label{sec:end2end}

We benchmark five models including BERT-Base, BERT-Large, GPT2, LLaMA, T5 and ViT. Among them, BERT and ViT are encoder-only, GPT2 and LLaMA are decoder-only, whereas T5 contains both encoder and decoder. 
We adopt the Bigbird mask and conduct experiments under three distinct settings of (batch size, sequence length): (1, 128), (8, 512), and (16, 2,048). Figure~\ref{fig:eva-EE-4090-A100} presents the end-to-end performance of the methods normalized to that of PyTorch Native on RTX 4090 and A100 GPUs. The missing bars indicate OOM for MCFuser or unsupported sequence length for ByteTransformer. As seen, STOF consistently delivers the highest speedups across the majority of models and settings on both GPU platforms. Even compared to the state-of-the-art PyTorch Compile, STOF achieves an average speedup of 1.3$\times$ and 1.4$\times$ on RTX 4090 and A100 GPUs, respectively. In addition to customizing the MHA kernel, the performance gain of STOF also comes from operator fusion and parameter tuning.

For the setting (16, 2,048), STOF achieves 1.5$\times$, 1.5$\times$, 1.2$\times$, 1.3$\times$, 1.1$\times$, and 1.2$\times$ speedups over PyTorch Compile for the six models on RTX 4090 GPU. A similar trend can be observed on A100 GPU. The results indicate that the advantages of STOF are particularly pronounced for larger input scales. The reason is attributed to the significant reduction in the absolute time of the bottleneck MHA computation. This demonstrates that STOF has the potential to be applied to future GPU generations with larger memory capacity.

\subsection{Tuning Cost}
\label{sec:search}

Table~\ref{tab:tuning_time} lists the tuning time of STOF, MCFuser, and Bolt for end-to-end inference on A100 GPU in seconds, where BERT-B/L is BERT-Base/Large. Note that PyTorch Native, PyTorch Compile, and ByteTransformer are not included due to the lack of tuning support. As seen, the tuning time of STOF is less than that of MCFuser and Bolt in all cases. This advantage becomes more prominent when the input scale is large. Since the tuning process of operator fusion module in STOF is positively correlated with the input tensor, the tuning cost per iteration increases moderately, but it does not grow linearly with respect to the overall tuning time. For the setting (16, 2,048), STOF is on average 6.7$\times$ and 6.9$\times$ faster than MCFuser and Bolt. This is mainly because reward-based sampling enables STOF to find high-performance settings in a shorter time. On the other hand, the caching mechanism ensures that the same parameter setting in each fusion scheme will not be executed repeatedly, which particularly saves tuning time in scenarios with large input scales.

\subsection{Ablation Study}
\label{sec:ablation}

\begin{figure}[htbp]
    \centering
    \includegraphics[width=\linewidth]{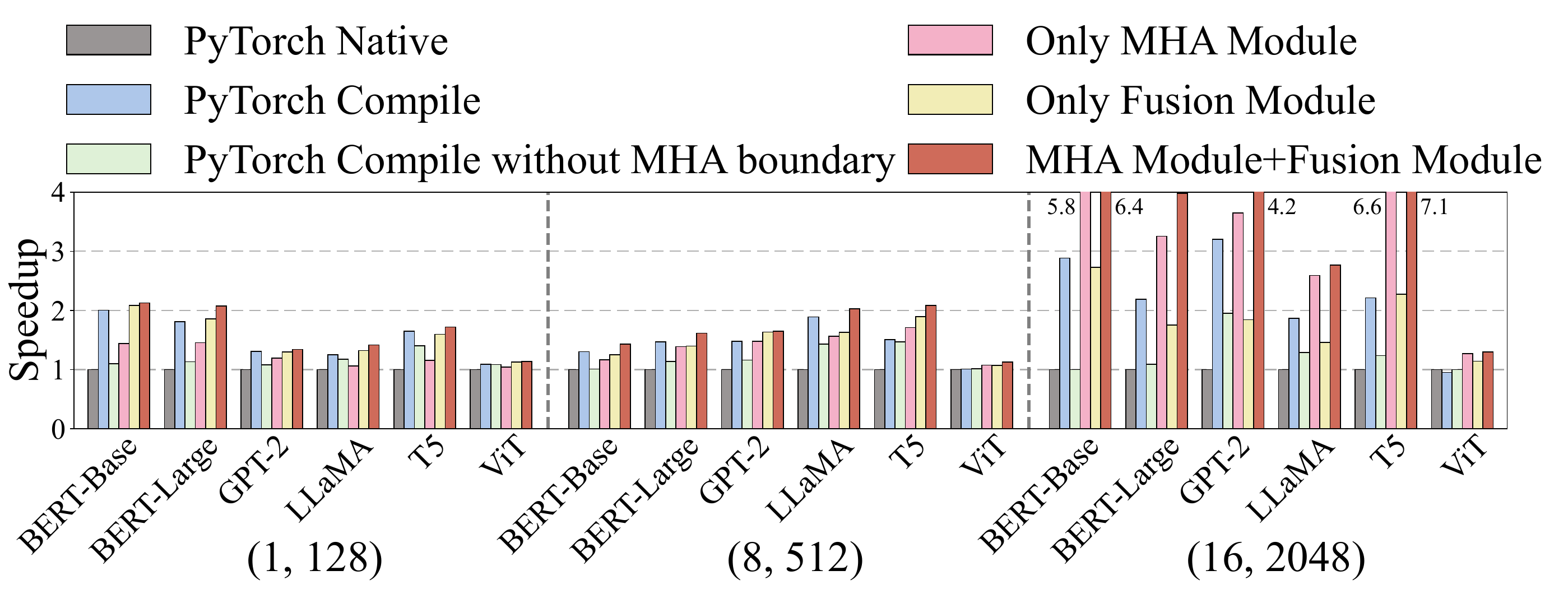}
    \caption{The speedup of STOF with only MHA module or only fusion module over PyTorch Native on A100 GPU.}
    \label{fig:5-ablation}
\end{figure}

Figure~\ref{fig:5-ablation} presents the speedup of STOF with only unified MHA module or only operator fusion module over PyTorch Native and PyTorch Compile on A100 GPU. 
For reference, the speedup of STOF with both modules is also shown in the figure. For PyTorch Compile, we also break the MHA boundary, transforming the whole computation graph into low-level meta operators for compilation optimization.

As seen, the operator fusion module contributes more to the performance when the input scale is small. Taking the setting of (1, 128) as an example, the speedup achieved by only fusion module is 19.5\% higher than that of only MHA module on average. In fact, the low sequence length and batch size lead to a small computational workload, which is particularly friendly to the fusion of CI operators. However, the contribution of the MHA module exceeds that of fusion module as the input scale increases. For the (16, 2,048) setting, the speedup of only MHA module is 2.0$\times$ on average, higher than that of only fusion module. Since MHA computation becomes the bottleneck, the high parallelism of the block-wise kernel is reflected in end-to-end inference. Note that STOF with both modules always achieves the highest speedup, indicating that the optimizations can complement each other. 
On the other hand, we find that breaking the MHA boundary would compromise these tailored kernel optimizations. The results show that such boundary breaking causes up to 1.5$\times$ slowdown compared to preserving it.



\subsection{Overhead Analysis}
\label{sec:overhead}

\begin{figure}[htbp]
    \centering
    \includegraphics[width=\linewidth]{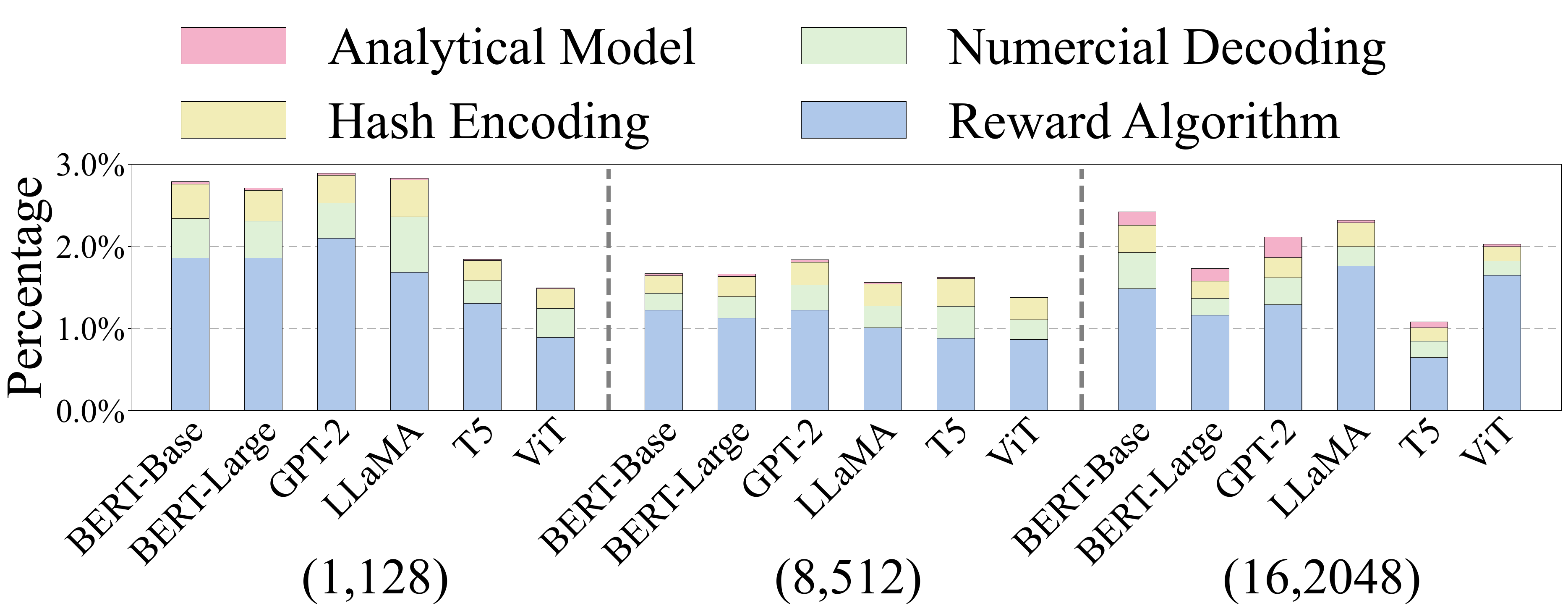}
    \caption{Time breakdown of the STOF overhead normalized to the tuning process on A100 GPU.}
    \label{fig:5-Percentage}
\end{figure}

The STOF overhead mainly includes the analysis model, scheme conversion (i.e., hash encoding and numerical decoding), and reward algorithm. The analysis model is reflected in MHA kernel selection and fusion scheme initialization. Figure~\ref{fig:5-Percentage} presents the time breakdown of STOF overhead normalized to the tuning process on A100 GPU. As seen, the time proportion of scheme conversion and reward algorithm is relatively smaller when the input scale is large. This is because these overheads are dominated by the model structure, and a larger input scale will lead to a longer tuning time, thus diluting this proportion. In contrast, the proportion of analytical model increases with the input scale. The primary reason is that the overhead for analyzing mask blocks increases with longer sequence lengths. Nevertheless, the analysis constitutes at most 0.5\% of the total time. Overall, STOF accounts for less than 3\% of the total tuning time, making it highly acceptable in the context of model fine-tuning.

\subsection{Discussion}
\label{sec:discussion}

\subsubsection{Newer GPU Architectures}

In addition to NVIDIA Ampere and Ada architectures, we have conducted preliminary tests on newer hopper architecture (i.e., NVIDIA H20 GPU). The results show that STOF consistently outperforms FA2, achieving up to 1.4$\times$ speedup for MHA computation. This proves that kernel optimizations of STOF are universal across GPU architectures. We plan to extend this evaluation to include FA3 for future work.


\subsubsection{Longer Sequence Lengths}

We explore sequence lengths ranging from 4k to 16k and batch size of 1 on NVIDIA A100 GPU. STOF achieves significant speedups over the SOTA PyTorch Compile, reaching 4.1$\times$, 11.1$\times$, and 16.8$\times$ at 4k, 8k, and 16k, respectively. In addition, all baselines except STOF encounter Out-of-Memory (OOM) errors at sequence length of 32k, whereas STOF reaches OOM at 64k. The results indicate that STOF exhibits greater performance improvement for ultra-long sequence lengths, as well as significantly saving GPU memory.


\subsubsection{Dynamic Mask Patterns}

STOF is inherently positioned to support dynamic mask patterns due to its flexible design. For example, MInference~\cite{jiang2024minference} could serve as a sophisticated frontend to discover dynamic patterns, with STOF as the execution backend. The main challenge lies in efficiently integrating MInference's offline pattern determination and online index generation into STOF's compilation pipeline with minor overhead. For future work, we plan to extend the analytical model to determine optimal configurations at runtime based on input token sequence.

\section{Related Work}  
\label{sec:relatedwork}

\textbf{Hardware Accelerators for Attention.} 
Recent works have considered the inherent parallelism and memory access patterns to design customized accelerators~\cite{ham20203,lu2020hardware,ham2021elsa,zhou2022transpim,fan2022adaptable,wang2023cta,you2023vitcod,qin2023fact,zhao2024hardware,bai2024swat,he2025papi,gu2025pim}. 
ELSA~\cite{ham2021elsa} utilizes an approximate similarity computation scheme to filter out insignificant relations.
ViTCoD~\cite{you2023vitcod} polarizes attention maps into denser and sparser patterns to reduce data movement.
He et al.~\cite{he2025papi} propose a PIM-enabled heterogeneous system that accelerates LLM decoding with a dynamic online scheduler.
This work focuses on attention optimizations on GPU, but has the potential to be applied to the emerging accelerators.

\noindent \textbf{Auto-tuning for Scientific Applications.} 
Existing works have designed auto-tuning approaches to handle the complexity 
of scientific applications~\cite{dongarra2018autotuning,pfaffe2019efficient,sun2021cstuner,sun2022gtuner,sun2023adaptive,randall2023transfer,cho2023harnessing,swann2024seer,xu2024enabling,wang2025accelerating}. Donggarra et al.~\cite{dongarra2018autotuning} perform batched calculation self-tuning on GPU for a series of numerically dense linear algebra operators.
Randall et al.~\cite{randall2023transfer} propose a generative method that achieves automatic adjustment based on few-shot transfer-learning. 
Plasticine~\cite{wang2025accelerating} introduces multi-level stencil representations and selects the better fusion strategy of stencil operators with a CNN-GNN-based model.
The above works provide references for the implementation of this paper.

\section{Conclusion}
\label{sec:conclusion}

In this paper, we propose STOF, an efficient framework with flexible masking and operator fusion for optimizing sparse Transformer on GPU. First, we propose a unified MHA module that implements row-wise and block-wise kernels with unique storage formats and optimizations. Then, we propose an operator fusion module that enables fusion expansion and parameter tuning as well as mapping the fusion schemes to compilation templates. The experimental results show that STOF outperforms the state-of-the-art works in terms of MHA computation and end-to-end inference.
For future work, we plan to extend STOF to support PaddlePaddle\footnote{https://github.com/PaddlePaddle/Paddle} and to incorporate it transparently into the compiler stack.



\section*{Acknowledgements}
We sincerely thank our shepherd, Gagan Agrawal, and the anonymous reviewers for their insightful feedback that greatly improved this paper. This work is supported by National Natural Science Foundation of China (Grant No. 62402525, 62322201, U23B2020, 62402526), Beijing Natural Science Foundation (Grant No. 4244086), and CCF-Baidu Open Fund. Qingxiao Sun is the corresponding author.


\bibliography{bibliography}

@inproceedings{dai2025STOF-AE,
  title={PPoPP26\_AE\_STOF\_CODE},
  author={Wenhao Dai},
  howpublished = {\url{https://doi.org/10.5281/zenodo.17705801}},
  year={2025}
}

@inproceedings{dosovitskiy2021vit,
  title={An image is worth 16x16 words: Transformers for image recognition at scale},
  author={Dosovitskiy, Alexey and Beyer, Lucas and Kolesnikov, Alexander and Weissenborn, Dirk and Zhai, Xiaohua and Unterthiner, Thomas and Dehghani, Mostafa and Minderer, Matthias and Heigold, Georg and Gelly, Sylvain and others},
  booktitle={International Conference on Learning Representations (ICLR '21)},
  year={2021}
}

@inproceedings{fan2025spinfer,
  title={Spinfer: Leveraging low-level sparsity for efficient large language model inference on gpus},
  author={Fan, Ruibo and Yu, Xiangrui and Dong, Peijie and Li, Zeyu and Gong, Gu and Wang, Qiang and Wang, Wei and Chu, Xiaowen},
  booktitle={Proceedings of the Twentieth European Conference on Computer Systems (EuroSys '25)},
  year={2025}
}

@article{lin2022survey,
  title={A survey of Transformers},
  author={Lin, Tianyang and Wang, Yuxin and Liu, Xiangyang and Qiu, Xipeng},
  journal={AI open},
  volume={3},
  pages={111--132},
  year={2022}
}

@inproceedings{cheng2025pipethreader,
  title={PipeThreader: Software-Defined Pipelining for Efficient DNN Execution},
  author={Cheng, Yu and Wang, Lei and Shi, Yining and Xia, Yuqing and Ma, Lingxiao and Xue, Jilong and Wang, Yang and Mo, Zhiwen and Chen, Feiyang and Yang, Fan and others},
  booktitle={19th USENIX Symposium on Operating Systems Design and Implementation (OSDI '25)},
  year={2025}
}

@inproceedings{tillet2019triton,
  title={Triton: An intermediate language and compiler for tiled neural network computationsf},
  author={Tillet, Philippe and Kung, Hsiang-Tsung and Cox, David},
  booktitle={ACM SIGPLAN International Workshop on Machine Learning and Programming Languages},
  year={2019},
}

@article{wang2024raptor,
  title = {Raptor-T: A fused and memory-efficient sparse Transformer for long and variable-length sequences},
  author = {Wang, Hulin and Yang, Donglin and Xia, Yaqi and Zhang, Zheng and Wang, Qigang and Fan, Jianping and Zhou, Xiaobo and Cheng, Dazhao},
  journal={ IEEE Transactions on Computers },
  volume = {73},
  number = {7},
  pages = {1852–1865},
  year = {2024},
}

@article{radford2019language,
  title={Language models are unsupervised multitask learners},
  author={Radford, Alec and Wu, Jeffrey and Child, Rewon and Luan, David and Amodei, Dario and Sutskever, Ilya and others},
  journal={OpenAI blog},
  volume={1},
  number={8},
  pages={9},
  year={2019}
}

@article{Raffel2020T5,
  author  = {Raffel, Colin and Shazeer, Noam and Roberts, Adam and Lee, Katherine and Narang, Sharan and MMatena, ichael and Zhou, Yanqi and Li, Wei and Liu, Peter J. },
  title   = {Exploring the limits of transfer learning with a unified text-to-text Transformer},
  journal = {Journal of Machine Learning Research},
  volume  = {21},
  number  = {140},
  pages   = {1-67},
  year    = {2020}
}

@inproceedings{ansel2024pytorch,
  title={PyTorch 2: Faster machine learning through dynamic python bytecode transformation and graph compilation},
  author={Ansel, Jason and Yang, Edward and He, Horace and Gimelshein, Natalia and Jain, Animesh and Voznesensky, Michael and Bao, Bin and Bell, Peter and Berard, David and Burovski, Evgeni and others},
  booktitle = {International Conference on Architectural Support for Programming Languages and Operating Systems (ASPLOS '24)},
  year={2024},
}

@inproceedings{kao2023flat,
  title={FLAT: An optimized dataflow for mitigating attention bottlenecks},
  author={Kao, Sheng-Chun and Subramanian, Suvinay and Agrawal, Gaurav and Yazdanbakhsh, Amir and Krishna, Tushar},
  booktitle = {International Conference on Architectural Support for Programming Languages and Operating Systems (ASPLOS '23)},
  year={2023},
}

@inproceedings{adnan2024keyformer,
  title={Keyformer: KV cache reduction through key tokens selection for efficient generative inference},
  author={Adnan, Muhammad and Arunkumar, Akhil and Jain, Gaurav and Nair, Prashant and Soloveychik, Ilya and Kamath, Purushotham},
  booktitle = {Conference on Machine Learning and Systems (MLSys '24)},
  year={2024},
}

@article{beltagy2020longformer,
  title={Longformer: The long-document Transformer},
  author={Beltagy, Iz and Peters, Matthew E and Cohan, Arman},
  journal={arXiv preprint arXiv:2004.05150},
  year={2020}
}

@inproceedings{zaheer2020big,
    title = {Big Bird: Transformers for longer sequences},
    author = {Zaheer, Manzil and Guruganesh, Guru and Dubey, Avinava and Ainslie, Joshua and Alberti, Chris and Ontanon, Santiago and Pham, Philip and Ravula, Anirudh and Wang, Qifan and Yang, Li and Ahmed, Amr},
    booktitle = {Conference on Neural Information Processing Systems (NeurIPS '20)},
    year = {2020},
}

@inproceedings{vaswani2017A,
  title={Attention is all you need},
  author={Vaswani, Ashish and Shazeer, Noam and Parmar, Niki and Uszkoreit, Jakob and Jones, Llion and Gomez, Aidan N and Kaiser, {\L}ukasz and Polosukhin, Illia},
  booktitle = {Conference on Neural Information Processing Systems (NeurIPS '17)},
  year={2017},
}

@inproceedings{devlin2019bert,
  title={BERT: Pre-training of deep bidirectional Transformers for language understanding},
  author={Devlin, Jacob and Chang, Ming-Wei and Lee, Kenton and Toutanova, Kristina},
  booktitle={Annual Conference of the North American chapter of the association for computational linguistics: human language technologies},
  year={2019}
}

@article{child2019generating,
  title={Generating long sequences with sparse Transformers},
  author={Child, Rewon and Gray, Scott and Radford, Alec and Sutskever, Ilya},
  journal={arXiv preprint arXiv:1904.10509},
  year={2019}
}

@article{roy2021efficient,
  title={Efficient content-based sparse attention with routing Transformers},
  author={Roy, Aurko and Saffar, Mohammad and Vaswani, Ashish and Grangier, David},
  journal={Transactions of the Association for Computational Linguistics},
  volume={9},
  number={},
  pages={53--68},
  year={2021},
}

@inproceedings{liu2023deja,
  title={Deja Vu: Contextual sparsity for efficient LLMs at inference time},
  author={Liu, Zichang and Wang, Jue and Dao, Tri and Zhou, Tianyi and Yuan, Binhang and Song, Zhao and Shrivastava, Anshumali and Zhang, Ce and Tian, Yuandong and Re, Christopher and others},
  booktitle={International Conference on Machine Learning (ICML '23)},
  year={2023},
}

@inproceedings{chen2018tvm,
  title={TVM: An automated end-to-end optimizing compiler for deep learning},
  author={Chen, Tianqi and Moreau, Thierry and Jiang, Ziheng and Zheng, Lianmin and Yan, Eddie and Shen, Haichen and Cowan, Meghan and Wang, Leyuan and Hu, Yuwei and Ceze, Luis and others},
  booktitle={USENIX Symposium on Operating Systems Design and Implementations (OSDI '18)},
  year={2018},
}

@article{ragan2013halide,
  title={Halide: A language and compiler for optimizing parallelism, locality, and recomputation in image processing pipelines},
  author={Ragan-Kelley, Jonathan and Barnes, Connelly and Adams, Andrew and Paris, Sylvain and Durand, Fr{\'e}do and Amarasinghe, Saman},
  journal={ACM Sigplan Notices},
  volume={48},
  number={6},
  pages={519--530},
  year={2013},
}

@inproceedings{niu2021dnnfusion,
  title={DNNFusion: Accelerating deep neural networks execution with advanced operator fusion},
  author={Niu, Wei and Guan, Jiexiong and Wang, Yanzhi and Agrawal, Gagan and Ren, Bin},
  booktitle={International Conference on Programming Language Design and Implementation (PLDI '21)},
  year={2021},
}

@inproceedings{shi2023welder,
  title={Welder: Scheduling deep learning memory access via tile-graph},
  author={Shi, Yining and Yang, Zhi and Xue, Jilong and Ma, Lingxiao and Xia, Yuqing and Miao, Ziming and Guo, Yuxiao and Yang, Fan and Zhou, Lidong},
  booktitle={USENIX Symposium on Operating Systems Design and Implementations (OSDI '23)},
  year={2023},
}

@inproceedings{dao2022flashattention,
  title={Flash{A}ttention: Fast and Memory-Efficient Exact Attention with {IO}-Awareness},
  author={Dao, Tri and Fu, Daniel Y. and Ermon, Stefano and Rudra, Atri and R{\'e}, Christopher},
  booktitle={Advances in Neural Information Processing Systems (NeurIPS '22)},
  year={2022}
}

@inproceedings{dao2023flashattention2,
  title={Flash{A}ttention-2: Faster Attention with Better Parallelism and Work Partitioning},
  author={Dao, Tri},
  booktitle={International Conference on Learning Representations (ICLR '23)},
  year={2024}
}

@article{shah2024flashattention3,
  title={FlashAttention-3: Fast and accurate attention with asynchrony and low-precision},
  author={Shah, Jay and Bikshandi, Ganesh and Zhang, Ying and Thakkar, Vijay and Ramani, Pradeep and Dao, Tri},
  journal={Advances in Neural Information Processing Systems},
  volume={37},
  pages={68658--68685},
  year={2024}
}

@inproceedings{fang2021turboTransformers,
  title={TurboTransformers: An efficient GPU serving system for Transformer models},
  author={Fang, Jiarui and Yu, Yang and Zhao, Chengduo and Zhou, Jie},
  booktitle={ACM SIGPLAN Symposium on Principles and Practice of Parallel Programming (PPoPP '21)},
  year={2021},
}

@inproceedings{zhai2023byteTransformer,
  title={ByteTransformer: A high-performance Transformer boosted for variable-length inputs}, 
  author={Zhai, Yujia and Jiang, Chengquan and Wang, Leyuan and Jia, Xiaoying and Zhang, Shang and Chen, Zizhong and Liu, Xin and Zhu, Yibo},
  booktitle={International Parallel \& Distributed Processing Symposium (IPDPS '23)},
  year={2023},
}

@inproceedings{aminabadi2022deepspeed,
  title={DeepSpeed Inference: Enabling efficient inference of Transformer models at unprecedented scale},
  author={Aminabadi, Reza Yazdani and Rajbhandari, Samyam and Awan, Ammar Ahmad and Li, Cheng and Li, Du and Zheng, Elton and Ruwase, Olatunji and Smith, Shaden and Zhang, Minjia and Rasley, Jeff and others},
  booktitle={International Conference on High Performance Computing, Networking, Storage and Analysis (SC '22)},
  year={2022},
}

@inproceedings{wang2022lightseq2,
  title={LightSeq2: Accelerated training for Transformer-based models on GPUs},
  author={Wang, Xiaohui and Wei, Yang and Xiong, Ying and Huang, Guyue and Qian, Xian and Ding, Yufei and Wang, Mingxuan and Li, Lei},
  booktitle={International Conference on High Performance Computing, Networking, Storage and Analysis (SC '22)},
  year={2022},
}

@inproceedings{zheng2022astitch,
  title={AStitch: Enabling a new multi-dimensional optimization space for memory-intensive ML training and inference on modern SIMT architectures},
  author={Zheng, Zhen and Yang, Xuanda and Zhao, Pengzhan and Long, Guoping and Zhu, Kai and Zhu, Feiwen and Zhao, Wenyi and Liu, Xiaoyong and Yang, Jun and Zhai, Jidong and others},
  booktitle={International Conference on Architectural Support for Programming Languages and Operating Systems (ASPLOS '18)},
  year={2022},
}

@inproceedings{zheng2023chimera,
  title     = {Chimera: An analytical optimizing framework for effective compute-intensive operators fusion},
  author    = {Zheng, Size and Chen, Siyuan and Song, Peidi and Chen, Renze and Li, Xiuhong and Yan, Shengen and Lin, Dahua and Leng, Jingwen and Liang, Yun},
  booktitle = {Proceedings of the 29th IEEE International Symposium on High Performance Computer Architecture (HPCA '23)},
  year      = {2023},
}

@inproceedings{zheng2020flextensor,
  title={FlexTensor: An automatic schedule exploration and optimization framework for tensor computation on heterogeneous system},
  author={Zheng, Size and Liang, Yun and Wang, Shuo and Chen, Renze and Sheng, Kaiwen},
  booktitle={International Conference on Architectural Support for Programming Languages and Operating Systems (ASPLOS '20)},
  year={2020},
}

@inproceedings{sun2021cstuner,
  title={csTuner: Scalable auto-tuning framework for complex stencil computation on GPUs},
  author={Sun, Qingxiao and Liu, Yi and Yang, Hailong and Jiang, Zhonghui and Liu, Xiaoyan and Dun, Ming and Luan, Zhongzhi and Qian, Depei},
  booktitle={IEEE International Conference on Cluster Computing (CLUSTER '21)},
  year={2021},
}

@inproceedings{zheng2020ansor,
  title={Ansor: Generating high-performance tensor programs for deep learning},
  author={Zheng, Lianmin and Jia, Chengfan and Sun, Minmin and Wu, Zhao and Yu, Cody Hao and Haj-Ali, Ameer and Wang, Yida and Yang, Jun and Zhuo, Danyang and Sen, Koushik and others},
  booktitle = {USENIX Symposium on Operating Systems Design and Implementation (OSDI '20)},
  year={2020},
}

@inproceedings{zheng2021tenset,
  title={TenSet: A large-scale program performance dataset for learned tensor compilers},
  author={Zheng, Lianmin and Liu, Ruochen and Shao, Junru and Chen, Tianqi and Gonzalez, Joseph E and Stoica, Ion and Ali, Ameer Haj},
  booktitle={Conference on Neural Information Processing Systems (NeurIPS '21)},
  year={2021},

}

@inproceedings{ma2020rammer,
  title={Rammer: Enabling holistic deep learning compiler optimizations with rTasks},
  author={Ma, Lingxiao and Xie, Zhiqiang and Yang, Zhi and Xue, Jilong and Miao, Youshan and Cui, Wei and Hu, Wenxiang and Yang, Fan and Zhang, Lintao and Zhou, Lidong},
  booktitle = {USENIX Symposium on Operating Systems Design and Implementation (OSDI '20)},
  year={2020},
}

@inproceedings{pfaffe2019efficient,
  title={Efficient hierarchical online-autotuning: A case study on polyhedral accelerator mapping},
  author={Pfaffe, Philip and Grosser, Tobias and Tillmann, Martin},
  booktitle={International Conference on Supercomputing (ICS '19)},
  pages = {354–366},
  year={2019},
  organization={ACM}
}

@article{dongarra2018autotuning,
  title={Autotuning numerical dense linear algebra for batched computation with GPU hardware accelerators},
  author={Dongarra, Jack and Gates, Mark and Kurzak, Jakub and Luszczek, Piotr and Tsai, Yaohung M},
  journal={Proceedings of the IEEE},
  volume={106},
  number={11},
  pages={2040--2055},
  year={2018}
}

@article{touvron2023llama,
  title={LLaMA: Open and Efficient Foundation Language Models},
  author={Touvron, Hugo and Lavril, Thibaut and Izacard, Gautier and Martinet, Xavier and Lachaux, Marie-Anne and Lacroix, Timoth{\'e}e and Rozi{\`e}re, Baptiste and Goyal, Naman and Hambro, Eric and Azhar, Faisal and others},
  journal={arXiv preprint arXiv:2302.13971},
  year={2023}
}

@inproceedings{cho2023harnessing,
  title={Harnessing the crowd for autotuning high-performance computing applications},
  author={Cho, Younghyun and Demmel, James W and King, Jacob and Li, Xiaoye S and Liu, Yang and Luo, Hengrui},
  booktitle={International Parallel \& Distributed Processing Symposiumm (IPDPS '23)},
  year={2023},
}

@inproceedings{sun2022gtuner,
  title={GTuner: Tuning DNN computations on GPU via graph attention network},
  author={Sun, Qi and Zhang, Xinyun and Geng, Hao and Zhao, Yuxuan and Bai, Yang and Zheng, Haisheng and Yu, Bei},
  booktitle={Asia and South Pacific Design Automation Conference (ASP-DAC '22)},
  year={2022},

}

@article{sun2023adaptive,
  title={Adaptive auto-tuning framework for global exploration of stencil optimization on GPUs},
  author={Sun, Qingxiao and Liu, Yi and Yang, Hailong and Jiang, Zhonghui and Luan, Zhongzhi and Qian, Depei},
  journal={IEEE Transactions on Parallel and Distributed Systems},
  volume={35},
  number={1},
  pages={20-33},
  year={2024}
}

@inproceedings{randall2023transfer,
  title={Transfer-learning-based autotuning using Gaussian copula},
  author={Randall, Thomas and Koo, Jaehoon and Videau, Brice and Kruse, Michael and Wu, Xingfu and Hovland, Paul and Hall, Mary and Ge, Rong and Balaprakash, Prasanna},
  booktitle={International Conference on Supercomputing (ICS '23)},
  year={2023},
}

@inproceedings{ouyang2022training,
  title={Training language models to follow instructions with human feedback},
  author={Ouyang, Long and Wu, Jeffrey and Jiang, Xu and Almeida, Diogo and Wainwright, Carroll and Mishkin, Pamela and Zhang, Chong and Agarwal, Sandhini and Slama, Katarina and Ray, Alex and others},
  booktitle={Conference on Neural Information Processing Systems (NeurIPS '22)},
  year={2022},
}

@inproceedings{ham20203,
  title={$A^3$: Accelerating attention mechanisms in neural networks with approximation},
  author={Ham, Tae Jun and Jung, Sung Jun and Kim, Seonghak and Oh, Young H and Park, Yeonhong and Song, Yoonho and Park, Jung-Hun and Lee, Sanghee and Park, Kyoung and Lee, Jae W and others},
  booktitle={High Performance Computer Architecture (HPCA '20)},
  year={2020},
}

@inproceedings{lu2020hardware,
  title={Hardware accelerator for multi-head attention and position-wise feed-forward in the Transformer},
  author={Lu, Siyuan and Wang, Meiqi and Liang, Shuang and Lin, Jun and Wang, Zhongfeng},
  booktitle={International System-on-Chip Conference (SOCC '20)},
  year={2020},
}

@inproceedings{ham2021elsa,
  title={ELSA: Hardware-software co-design for efficient, lightweight self-attention mechanism in neural networks},
  author={Ham, Tae Jun and Lee, Yejin and Seo, Seong Hoon and Kim, Soosung and Choi, Hyunji and Jung, Sung Jun and Lee, Jae W},
  booktitle={International Symposium on Computer Architecture (ISCA '21)},
  year={2021},
}

@inproceedings{zhou2022transpim,
  title={TransPIM: A memory-based acceleration via software-hardware co-design for Transformer},
  author={Zhou, Minxuan and Xu, Weihong and Kang, Jaeyoung and Rosing, Tajana},
  booktitle={High Performance Computer Architecture (HPCA '22)},
  year={2022},
}

@inproceedings{fan2022adaptable,
  title={Adaptable butterfly accelerator for attention-based NNs via hardware and algorithm co-design},
  author={Fan, Hongxiang and Chau, Thomas and Venieris, Stylianos I and Lee, Royson and Kouris, Alexandros and Luk, Wayne and Lane, Nicholas D and Abdelfattah, Mohamed S},
  booktitle={IEEE/ACM International Symposium on Microarchitecture (MICRO '22)},
  year={2022},
}

@inproceedings{wang2023cta,
  title={CTA: Hardware-software co-design for compressed token attention mechanism},
  author={Wang, Haoran and Xu, Haobo and Wang, Ying and Han, Yinhe},
  booktitle={High Performance Computer Architecture (HPCA '23)},
  year={2023},
}

@inproceedings{you2023vitcod,
  title={ViTCoD: Vision Transformer acceleration via dedicated algorithm and accelerator co-design},
  author={You, Haoran and Sun, Zhanyi and Shi, Huihong and Yu, Zhongzhi and Zhao, Yang and Zhang, Yongan and Li, Chaojian and Li, Baopu and Lin, Yingyan},
  booktitle={High Performance Computer Architecture (HPCA '23)},
  year={2023},
}

@inproceedings{qin2023fact,
  title={FACT: FFN-attention co-optimized Transformer architecture with eager correlation prediction},
  author={Qin, Yubin and Wang, Yang and Deng, Dazheng and Zhao, Zhiren and Yang, Xiaolong and Liu, Leibo and Wei, Shaojun and Hu, Yang and Yin, Shouyi},
  booktitle={International Symposium on Computer Architecture (ISCA '23)},
  year={2023},
}

@article{zhao2024hardware,
  author={Zhao, Jieru and Zeng, Pai and Shen, Guan and Chen, Quan and Guo, Minyi},
  journal={IEEE Transactions on Computer-Aided Design of Integrated Circuits and Systems}, 
  title={Hardware–software co-design enabling static and dynamic sparse attention mechanisms}, 
  volume={43},
  number={9},
  pages={2783-2796},
  year={2024},
}

@inproceedings{bai2024swat,
  title={SWAT: Scalable and efficient window attention-based Transformers acceleration on FPGAs},
  author={Bai, Zhenyu and Dangi, Pranav and Li, Huize and Mitra, Tulika},
  booktitle={Design Automation Conference (DAC '24)},
  year={2024},
}

@inproceedings{swann2024seer,
  title={Seer: Predictive runtime kernel selection for irregular problems},
  author={Swann, Ryan and Osama, Muhammad and Sangaiah, Karthik and Mahmud, Jalal},
  booktitle={Code Generation and Optimization (CGO '24)},
  year={2024},
}

@article{xu2024enabling,
author = {Xu, Jiaming and Huang, Shan and Li, Jinhao and Huang, Guyue and Xie, Yuan and Wang, Yu and Dai, Guohao},
title = {Enabling efficient sparse multiplications on GPUs with heuristic adaptability},
journal = {IEEE Transactions on Computer-Aided Design of Integrated Circuits and Systems},
year = {2024},
pages = {1-1},
volume = {PP},
}

@inproceedings{zhang2024mcfuser,
  title={MCFuser: High-performance and rapid fusion of memory-bound compute-intensive operators},
  author={Zhang, Zheng and Yang, Donglin and Zhou, Xiaobo and Cheng, Dazhao},
  booktitle={International Conference for High Performance Computing, Networking, Storage and Analysis (SC '24)},
  year={2024},
}

@inproceedings{xu2023alt,
  title={ALT: Breaking the wall between data layout and loop optimizations for deep learning compilation},
  author={Xu, Zhiying and Xu, Jiafan and Peng, Hongding and Wang, Wei and Wang, Xiaoliang and Wan, Haoran and Dai, Haipeng and Xu, Yixu and Cheng, Hao and Wang, Kun and others},
  booktitle={European Conference on Computer Systems (EuroSys '23)},
  year={2023},
}

@inproceedings{li2024accelerated,
  title={Accelerated auto-tuning of GPU kernels for tensor computations},
  author={Li, Chendi and Xu, Yufan and Saravani, Sina Mahdipour and Sadayappan, Ponnuswamy},
  booktitle={International Conference on Supercomputing (ICS '24)},
  year={2024},
}

@article{chang2024survey,
  title={A survey on evaluation of large language models},
  author={Chang, Yupeng and Wang, Xu and Wang, Jindong and Wu, Yuan and Yang, Linyi and Zhu, Kaijie and Chen, Hao and Yi, Xiaoyuan and Wang, Cunxiang and Wang, Yidong and others},
  journal={ACM Transactions on Intelligent Systems and Technology},
  volume={15},
  number={3},
  pages={1--45},
  year={2024},
}

@article{zhao2023survey,
  title={A survey of large language models},
  author={Zhao, Wayne Xin and Zhou, Kun and Li, Junyi and Tang, Tianyi and Wang, Xiaolei and Hou, Yupeng and Min, Yingqian and Zhang, Beichen and Zhang, Junjie and Dong, Zican and others},
  journal={arXiv preprint arXiv:2303.18223},
  volume={1},
  number={2},
  year={2023}
}

@article{guo2025deepseek,
  title={DeepSeek-R1: Incentivizing reasoning capability in LLMs via reinforcement learning},
  author={Guo, Daya and Yang, Dejian and Zhang, Haowei and Song, Junxiao and Zhang, Ruoyu and Xu, Runxin and Zhu, Qihao and Ma, Shirong and Wang, Peiyi and Bi, Xiao and others},
  journal={arXiv preprint arXiv:2501.12948},
  year={2025}
}

@article{achiam2023gpt,
  title={GPT-4 technical report},
  author={Achiam, Josh and Adler, Steven and Agarwal, Sandhini and Ahmad, Lama and Akkaya, Ilge and Aleman, Florencia Leoni and Almeida, Diogo and Altenschmidt, Janko and Altman, Sam and Anadkat, Shyamal and others},
  journal={arXiv preprint arXiv:2303.08774},
  year={2023}
}

@inproceedings{wang2024flashmask,
  title={FlashMask: Efficient and Rich Mask Extension of FlashAttention},
  author={Wang, Guoxia and Zeng, Jinle and Xiao, Xiyuan and Wu, Siming and Yang, Jiabin and Zheng, Lujing and Chen, Zeyu and Bian, Jiang and Yu, Dianhai and Wang, Haifeng},
  booktitle={International Conference on Learning Representations (ICLR '24)},
  year={2024}
}

@article{dong2024flex,
  title={Flex Attention: A programming model for generating optimized attention kernels},
  author={Dong, Juechu and Feng, Boyuan and Guessous, Driss and Liang, Yanbo and He, Horace},
  journal={arXiv preprint arXiv:2412.05496},
  year={2024}
}

@article{zheng2021neoflow,
  title={NeoFlow: A flexible framework for enabling efficient compilation for high performance DNN training},
  author={Zheng, Size and Chen, Renze and Jin, Yicheng and Wei, Anjiang and Wu, Bingyang and Li, Xiuhong and Yan, Shengen and Liang, Yun},
  journal={IEEE Transactions on Parallel and Distributed Systems},
  volume={33},
  number={11},
  pages={3220--3232},
  year={2021},
  publisher={IEEE}
}

@article{ma2019paddlepaddle,
  title={PaddlePaddle: An open-source deep learning platform from industrial practice},
  author={Ma, Yanjun and Yu, Dianhai and Wu, Tian and Wang, Haifeng},
  journal={Frontiers of Data and Computing},
  volume={1},
  number={1},
  pages={105--115},
  year={2019}
}

@Misc{xFormers2022,
  author =       {Benjamin, Lefaudeux and Francisco, Massa and Diana, Liskovich and Wenhan, Xiong and Vittorio, Caggiano and Sean, Naren and Min, Xu and Jieru, Hu and Marta, Tintore and Susan, Zhang and Patrick, Labatut and Daniel, Haziza and Luca, Wehrstedt and Jeremy, Reizenstein and Grigory, Sizov},
  title =        {xFormers: A modular and hackable Transformer modelling library},
  howpublished = {\url{https://github.com/facebookresearch/xformers}},
  year =         {2022}
}

@Misc{nvidia22faster,
  author =       {NVIDIA},
  howpublished = {https://github.com/NVIDIA/FasterTransformer},
  year =         {2022}
}

@Misc{nvidia22cutlass,
  author =       {NVIDIA},
  howpublished = {https://github.com/NVIDIA/cutlass},
  year =         {2022}
}

@inproceedings{zhao2021akg,
  title={AKG: Automatic kernel generation for neural processing units using polyhedral transformations},
  author={Zhao, Jie and Li, Bojie and Nie, Wang and Geng, Zhen and Zhang, Renwei and Gao, Xiong and Cheng, Bin and Wu, Chen and Cheng, Yun and Li, Zheng and others},
  booktitle={ACM SIGPLAN Conference on Programming Language Design and Implementation (PLDI '21)},
  year={2021},
}

@inproceedings{li2023exploiting,
  title={Exploiting subgraph similarities for efficient auto-tuning of tensor programs},
  author={Li, Mingzhen and Yang, Hailong and Zhang, Shanjun and Yu, Fengwei and Gong, Ruihao and Liu, Yi and Luan, Zhongzhi and Qian, Depei},
  booktitle={International Conference on Parallel Processing (ICPP '23)},
  year={2023},
}

@article{xing2022bolt,
  title={Bolt: Bridging the gap between auto-tuners and hardware-native performance},
  author={Xing, Jiarong and Wang, Leyuan and Zhang, Shang and Chen, Jack and Chen, Ang and Zhu, Yibo},
  journal={Proceedings of Machine Learning and Systems},
  volume={4},
  pages={204--216},
  year={2022}
}

@inproceedings{chen2024evt,
  title={EVT: Accelerating deep learning training with  Epilogue Visitor Tree},
  author={Chen, Zhaodong and Kerr, Andrew and Cai, Richard and Kosaian, Jack and Wu, Haicheng and Ding, Yufei and Xie, Yuan},
  booktitle={International Conference on Architectural Support for Programming Languages and Operating Systems (ASPLOS '24)},
  year={2024},
}

@inproceedings{niu2022tilespgemm,
  title={TileSpGEMM: A tiled algorithm for parallel sparse general matrix-matrix multiplication on GPUs},
  author={Niu, Yuyao and Lu, Zhengyang and Ji, Haonan and Song, Shuhui and Jin, Zhou and Liu, Weifeng},
  booktitle={ACM SIGPLAN Symposium on Principles and Practice of Parallel Programming (PPoPP '22)},
  year={2022},
}

@inproceedings{gu2025pim,
  title={PIM is all you need: A CXL-enabled GPU-free system for large language model inference},
  author={Gu, Yufeng and Khadem, Alireza and Umesh, Sumanth and Liang, Ning and Servot, Xavier and Mutlu, Onur and Iyer, Ravi and Das, Reetuparna},
  booktitle={Proceedings of the 30th ACM International Conference on Architectural Support for Programming Languages and Operating Systems (ASPLOS '25)},
  year={2025},
}

@inproceedings{he2025papi,
  title={Papi: Exploiting dynamic parallelism in large language model decoding with a processing-in-memory-enabled computing system},
  author={He, Yintao and Mao, Haiyu and Giannoula, Christina and Sadrosadati, Mohammad and G{\'o}mez-Luna, Juan and Li, Huawei and Li, Xiaowei and Wang, Ying and Mutlu, Onur},
  booktitle={Proceedings of the 30th ACM International Conference on Architectural Support for Programming Languages and Operating Systems (ASPLOS '25)},
  year={2025},
}

@inproceedings{wang2025accelerating,
  title={Accelerating Complex Stencil Computations with Adaptive Fusion Strategy},
  author={Wang, Siqi and Yang, Hailong and Wang, Pengbo and Du, Shaokang and Xu, Yufan and Sun, Qingxiao and Liu, Xiaoyan and Wang, Xuezhu and Liang, Xuning and Luan, Zhongzhi and others},
  booktitle={Proceedings of the 39th ACM International Conference on Supercomputing (ICS '25)},
  pages={265--278},
  year={2025}
}

@article{dosovitskiy2020image,
  title={An image is worth 16x16 words: Transformers for image recognition at scale},
  author={Dosovitskiy, Alexey and Beyer, Lucas and Kolesnikov, Alexander and Weissenborn, Dirk and Zhai, Xiaohua and Unterthiner, Thomas and Dehghani, Mostafa and Minderer, Matthias and Heigold, Georg and Gelly, Sylvain and others},
  journal={arXiv preprint arXiv:2010.11929},
  year={2020}
}

@inproceedings{jiang2024minference,
  title={Minference 1.0: Accelerating pre-filling for long-context llms via dynamic sparse attention},
  author={Jiang, Huiqiang and Li, Yucheng and Zhang, Chengruidong and Wu, Qianhui and Luo, Xufang and Ahn, Surin and Han, Zhenhua and Abdi, Amir H and Li, Dongsheng and Lin, Chin-Yew and others},
  booktitle={Conference on Neural Information Processing Systems (NeurIPS '24)},
  year={2024}
}

@inproceedings{gupta2025splat,
  title={SPLAT: A framework for optimised GPU code-generation for SParse reguLar ATtention},
  author={Gupta, Ahan and Yuan, Yueming and Jain, Devansh and Ge, Yuhao and Aponte, David and Zhou, Yanqi and Mendis, Charith},
  booktitle={Proceedings of the ACM on Programming Languages (OOPSLA '25)},
  year={2025},
}

@article{cai2025survey,
  title={A survey on mixture of experts in large language models},
  author={Cai, Weilin and Jiang, Juyong and Wang, Fan and Tang, Jing and Kim, Sunghun and Huang, Jiayi},
  journal={IEEE Transactions on Knowledge and Data Engineering},
  year={2025},
  publisher={IEEE}
}


\end{document}